\documentclass[10pt,twocolumn,letterpaper]{article}

\usepackage[pagenumbers]{cvpr} 
\usepackage{color,xcolor}
\usepackage{graphicx}
\usepackage{subfloat} 
\usepackage{amsmath}
\usepackage{amssymb}
\usepackage{booktabs}
\usepackage{subcaption}
\usepackage{multirow}
\usepackage{bm}
\usepackage{ifthen}
\usepackage{siunitx}
\usepackage{marvosym}
\usepackage{enumerate}
\usepackage{bbm}

\usepackage[pagebackref,breaklinks,colorlinks]{hyperref}

\usepackage[capitalize]{cleveref}
\crefname{section}{Sec.}{Secs.}
\Crefname{section}{Section}{Sections}
\Crefname{table}{Table}{Tables}
\crefname{table}{Tab.}{Tabs.}
\newcommand{\db}{\si{\deci\bel}}

\begin{document}

\title{Reflash Dropout in Image Super-Resolution}

\author{Xiangtao Kong$^{1,2,4*}$ \quad Xina Liu$^{1,2}$\thanks{Equal contributions} \quad Jinjin Gu$^{3,1,4}$ \quad Yu Qiao$^{1,4}$ \quad Chao Dong$^{1,4}$ \thanks{Corresponding author (e-mail: chao.dong@siat.ac.cn)}\\
$^{1}$ShenZhen Key Lab of Computer Vision and Pattern Recognition, SIAT-SenseTime Joint Lab,\\
Shenzhen Institutes of Advanced Technology, Chinese Academy of Sciences\\
$^{2}$University of Chinese Academy of Sciences\\
$^{3}$The University of Sydney\quad
$^{4}$Shanghai AI Laboratory, Shanghai, China\\
{\tt\small \{xt.kong, xn.liu, yu.qiao, chao.dong\}@siat.ac.cn, jinjin.gu@sydney.edu.au}
}
\maketitle

\begin{abstract}

Dropout is designed to relieve the overfitting problem in high-level vision tasks but is rarely applied in low-level vision tasks, like image super-resolution (SR).
As a classic regression problem, SR exhibits a different behaviour as high-level tasks and is sensitive to the dropout operation.
However, in this paper, we show that appropriate usage of dropout benefits SR networks and improves the generalization ability.
Specifically, dropout is better embedded at the end of the network and is significantly helpful for the multi-degradation settings.
This discovery breaks our common sense and inspires us to explore its working mechanism.
We further use two analysis tools -- one is from a recent network interpretation work, and the other is specially designed for this task.
The analysis results provide side proofs to our experimental findings and show us a new perspective to understand SR networks. 

\end{abstract}

\section{Introduction}
\label{sec:Introduction}
Image super-resolution (SR) is a classic low-level vision task aiming at restoring a high-resolution image from a low-resolution input.
Benefiting from the powerful convolutional neural networks (CNNs), deep SR networks~\cite{SRCNN,VDSR,SRResNet,EDSR,PAN,RDN,RCAN,SAN,kong2021classsr,chen2021attention} can easily fit the training data and achieve impressive results in a synthetic environment.
To further extend their success to real-world images, researchers begin to design blind SR methods \cite{liu2021blind}, which can deal with unknown downsampling kernels or degradations.
Recent advances have made significant progress by enriching the data diversity~\cite{wang2021realesrgan,zhang2021designing,feng2019suppressing,yoo2020rethinking} and enlarging the model capacity~\cite{wang2021unsupervised,luo2020unfolding}, but none of them has tried to improve the training strategy.
The overfitting problem will become prominent when the network scale increases significantly, resulting in a weak generalization ability.
Then what kind of training strategy is suitable for the blind SR task?
A simple yet surprising answer comes to our mind.
It is dropout~\cite{hinton2012improving}, which is originally designed to avoid overfitting and has been proved effective in high-level vision tasks.
In this work, we will dive into the usage of dropout and reflash it in super-resolution.

\begin{figure}[t!]
  \centering
  \includegraphics[height=3.5cm]{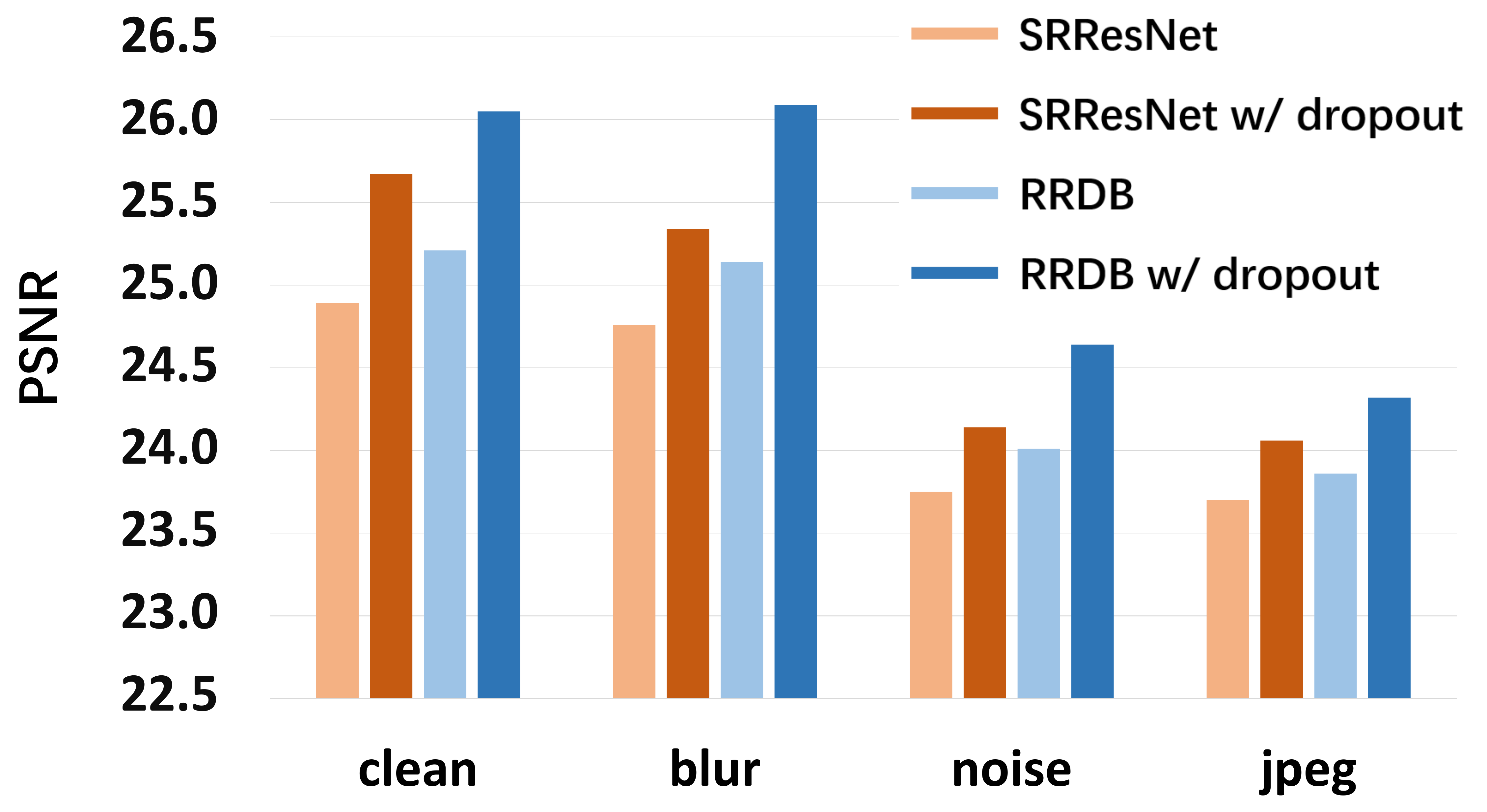}
  \vskip -0cm
  \caption{Dropout can significantly improve the performance of the models under the multi-degradation setting. It can even help SRResNet outperform RRDB, while the latter has ten times more parameters.
  There are the PSNR ($\db$) results of $\times 4$ SR models on Set5 with different degradations. For example, clean means input LR images without any degradations, noise means input LR images with noise. 
  }
  \label{fig:intro}
  \vskip -0.5cm
\end{figure}

Dropout seems to be in conflict with SR in nature.
Specifically, the mechanism of dropout is to disable some units and produce a number of sub-networks randomly.
Each sub-network is able to give an acceptable result.
However, SR is a standard regression problem, where network features and channels all have contributions to the final output.
If we randomly discard some features or pixels, the output performance will drop severely.
That is why we cannot see the application of dropout in SR, as well as other low-level vision tasks.
From another perspective, overfitting is not a severe problem in conventional SR tasks; thus, SR does not need dropout as well.
However, this situation changes nowadays.
First, overfitting has become a dominant problem for blind SR \cite{liu2021blind}.
Simply increasing the data and network scale cannot continuously improve the generalization ability.
Second, we have obtained a series of analysis tools in the area of network interpretation, assisting us in finding better ways of application.

To study dropout, we begin with its usage in the conventional non-blind settings.
After trying different dropout strategies, we can conclude detailed guidance of using dropout in SR.
With appropriate usage of dropout, the performance of SR models can improve significantly in both in-distribution (seen in the training set) and out-distribution (unseen) data.
\figurename~\ref{fig:intro} shows the performance before and after dropout, where the most significant PSNR gap can reach 0.95 $\db$.
It is worth noting that dropout can help SRResNet even outperform RRDB, while the latter has ten times more parameters.
More importantly, adding dropout is only one line of code and has no sacrifice on computation cost.
The most appealing part of this paper does not lie in the experiments but in the following analysis.
We adopt two novel interpretation tools, \ie, channel saliency map and deep degradation representation~\cite{liu2021discovering}) to analyze the behaviour of dropout.
We find that dropout can equalize the importance of feature maps, which could inherently improve the generalization ability.
There are also some other interesting observations, which all support our experimental results.
We believe that these analyses can help us understand the working mechanism of SR networks and inspire more effective training strategies in the future.

\section{Related Work}
\label{sec:Related-Work}

\noindent\textbf{Super-Resolution.}\quad
CNN-based SR networks~\cite{SRCNN,VDSR,SRResNet,EDSR,RDN,PAN,RCAN,SAN,kong2021classsr,chen2021attention,gu2020image,chen2021attention} aim to reconstruct a high-resolution (HR) image from its low-resolution (LR) observation.
These networks are usually trained in a conventional SR setting where the LR images are produced by the bicubic downsampling.
However, overfitting to one degradation leads to poor performance in real-world scenarios.
Recently, several works have been proposed to handle multiple degradations and even unknown degradations.
Some methods try to first predict degradations explicitly or implicitly and then conditionally reconstruct according to the predicted degradation, \eg, IKC~\cite{IKC}, KernelGAN \cite{bell2019blind}, and DASR \cite{wang2021unsupervised}.
These approaches rely on a predefined limited degradation model and still cannot generalize to the data that the degradation model can not cover.
Some other methods try to learn end-to-end SR networks that can generalize to a large range of real-world data, \eg, RealESRGAN~\cite{wang2021realesrgan} and BSRGAN~\cite{zhang2021designing}.
These methods assume that training networks on diverse data can improve generalization capabilities and randomly generate a large amount of training data with different degradations during training.
But there is no discussion under which training strategy can maximize the generalization ability. These methods still use the most straightforward direct optimization strategy.

\vspace{2pt}
\noindent\textbf{Dropout.}\quad
Dropout is a regularization technique and is first proposed to address the overfitting problem in classification networks.
The key idea is to randomly drop units (along with their connections) from the neural network during training.
Therefore, in the training phase, dropout makes only part of the network to be updated each time, and it is an efficient method of averaging sub-networks. 
Dropout follows a long line of research.
A large number of variants have been developed~\cite{wan2013regularization,goodfellow2013maxout,larsson2017ultra,spatialdropout} to improve the use of dropout and to adopt dropout in different practical problems. 
Among them, two works are more relevant to our work.
SpatialDropout~\cite{spatialdropout} (channel-wise dropout) formulates a new dropout method to zero out channels from the feature map. 
When the input has a strong spatial correlation, this method performs better than previous dropout strategies. 
Different from the original method of adding dropout at the fully connected layers, DropBlock~\cite{ghiasi2018dropblock} applies dropout to residual blocks (behind convolution layer and skip connection) and then explores using dropout in different parts of networks.

Besides, to interpret the success of dropout, various works have attempted to analyze it from different perspectives \cite{helmbold2017surprising, jain2015drop,bouthillier2015dropout,gal2016dropout}.
Srivastava \etal~\cite{srivastava2014dropout} argue that the dropout method samples from an exponential number of different ``thinned'' networks and approximates the effect of averaging the predictions of all these thinned networks at test time.
Some other works attempt to theoretically study the generalization performance for the deep neural network with dropout. 
For instance, Gao \etal~\cite{gao2016dropout} point out that dropout can help to reduce the networks' Rademacher complexity.
However, most of these improvements, explanations and discussions are aimed at classification tasks.
Although dropout has been widely used in classification tasks, its role in super-resolution has not been explored.

\section{Observation}
\label{sec:Observation}

\begin{figure}[t!]
  \centering
  \includegraphics[width=\linewidth]{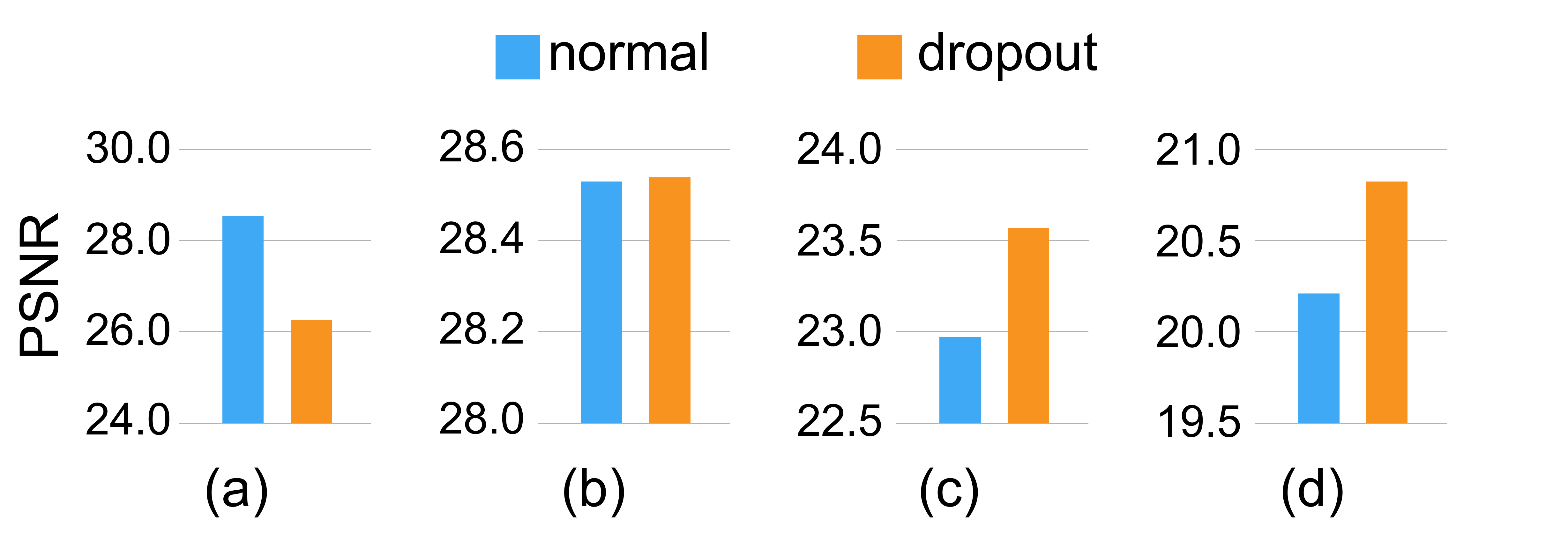}
  \vskip -0.3cm
  \caption{Performance of SRResNet with different settings on Manga109 with $\times 4$. 
  (a) Naive applying of dropout harms SR.
  (b) Appropriate applying of dropout does not affect SR.
  (c) (d) Dropout is beneficial for SR in some situations.
  }
  
  \label{fig:bar}
  \vskip -0.5cm
\end{figure}

\begin{figure*}[t!]
  \centering
  \includegraphics[width=0.9\linewidth]{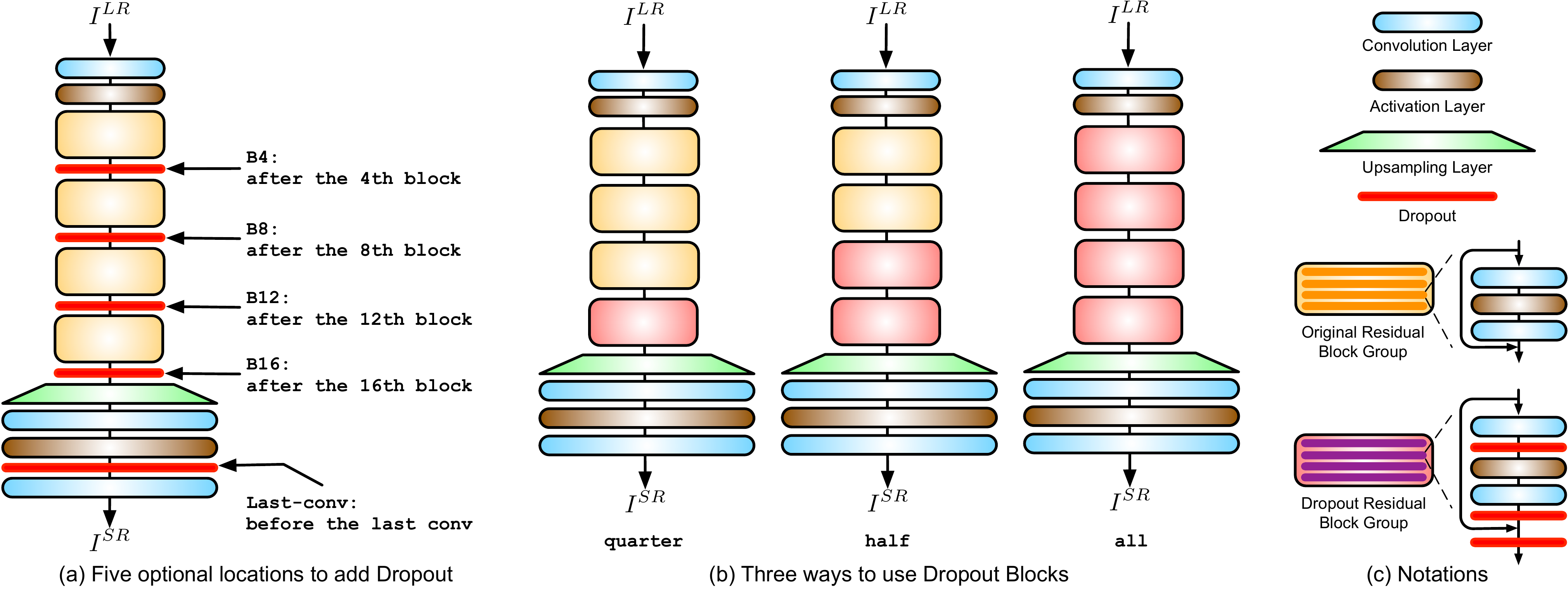} 
  \vskip -0.2cm
  \caption{Different ways to apply dropout in SRResNet: (a) illustrates five optional positions where we can add a single dropout layer, marked with red layers; (b) illustrates three ways to add dropout inside the residual blocks; (c) presents the notation.}
  \label{fig:arch}
  \vskip -0.4cm
\end{figure*}

We have made some primary attempts to adopt dropout in SR and find that the networks exhibit completely different behaviours under different settings.
It is hard to reach a consistent conclusion but will motivate the following study.

\vspace{2pt}
\noindent\textbf{Dropout is harmful for SR.}\quad
This experiment is conducted under the conventional SR setting, where the only degradation is the bicubic downsampling.
We adopt the widely-used dropout strategy -- channel-wise dropout~\cite{spatialdropout} (randomly zero out the entire channels) after each convolution layer of SRResNet~\cite{SRResNet}.
As expected, the performance drops dramatically (see \figurename~\ref{fig:bar}{\color[HTML]{FF0000}a}).
This result exactly conforms to our common sense.
It indicates that the regression problem is different from the classification problem.
In regression, each element in the network contributes to the final output, which is a continuous intensity value but not a discrete class label.
More experiments in Section~\ref{sec:How to Apply Dropout in SR Networks} show that most common dropout strategies in classification do not work well on SR.

\vspace{2pt}
\noindent\textbf{Dropout does not affect SR.}\quad
However, we find a special case that does not coincide with the above observation.
Under the same setting, we add channel-wise dropout only before the last convolution layer.
The final performance is not affected at all, see \figurename~\ref{fig:bar}{\color[HTML]{FF0000}b}.
This phenomenon is interesting.
It indicates that the features in that layer can be randomly masked, which does not influence the regression results.
We have also tried to discard a few features during testing, and found no apparent performance drop, see Section~\ref{sec:Channel Saliency Map}.
What happens to those features?
Does that mean the regression and classification networks have something in common?
This inspires our curiosity.

\vspace{2pt}
\noindent\textbf{Dropout is beneficial for SR.}\quad
The last observation is even more interesting.
We find that under the multiple-degradation setting, dropout can even benefit SR.
A simple experimental setting is as follows.
The training data contain enough degradations, namely Real-SRResNet.
We add dropout at the second last convolution layer.
The performance is tested on bicubic (seen in the training set) and nearest neighbour (unseen) downsampling dataset.
From \figurename~\ref{fig:bar}{\color[HTML]{FF0000}c} and ~\ref{fig:bar}{\color[HTML]{FF0000}d}, we can observe that dropout improves performance in both in-distribution and out-distribution data.
This indicates that dropout improves the generalization ability to some extent.
Does this finding have the same theoretical interpretation as the previous one?
Can we find other cases where dropout benefits SR?
All the observations above can provide us with a clue to recover the effectiveness of dropout in low-level tasks.
We will steadily go through this process by detailing the dropout strategies, describing the experiments and revealing the inner working mechanisms.


\section{Apply Dropout in SR Network}
\label{sec:Dropout}
To explore the application strategies of dropout, we borrow the successful experience from high-level vision tasks. 
In this section, we will systematically review the feasible implementations of dropout in previous works, and apply them in SR networks.
Our study is based on two representative SR networks -- SRResNet~\cite{SRResNet} and RRDB~\cite{ESRGAN}.
Our conclusion can be easily generalized to other CNN based SR networks~\cite{zhang2019ranksrgan,wang2021realesrgan,zhang2021designing}, which share similar architectures. 
As a simple and flexible operation, dropout has many application ways.
In general, the effect of dropout mainly depends on two aspects, one is the dropout position, and the other is the dropout strategy.
We will discuss them as follows.

\subsection{Dropout Position}
\label{sec:DropoutPosition}
We explore these potential positions for applying dropout in SR networks through analogy analysis with previous studies in high-level vision.
The positions can be mainly divided into three categories.
It is very helpful to refer to \figurename~\ref{fig:arch} when reading the following description:
\begin{enumerate}[(1)]
    \item Use dropout before the final output layer. Hinton \etal \cite{hinton2012improving} first introduce dropout and apply it at the fully connected layers before the final classification layer. Similarly, we also apply the dropout layer before the output convolutional layer (from the feature channels to the RGB channels). We use \texttt{last-conv} to represent this method.
    
    \item Use dropout at the middle of the network. Many works also try to use dropout at the middle of the network, \eg, after a special convolution layer \cite{spatialdropout} and at certain locations \cite{dropblock}. Without loss of generality, we split the SRResNet residual blocks (16 blocks) into four groups. Each group consists of four residual blocks. We choose \texttt{B4}, \texttt{B8}, \texttt{B12}, \texttt{B16} as representative positions, where the number indicates that dropout is added after which blocks.
    
    \item Use multiple dropout layers in a residual network. Ghiasi \etal \cite{dropblock} suggest that we can apply the dropout layer inside the residual block and use these ``dropped residual blocks'' multiple times. \figurename~\ref{fig:arch}{\color[HTML]{FF0000}c} shows the detail of the  ``dropped residual blocks''. According to their experiments, using this ``dropped residual blocks'' at the deep locations of the network could generate the best results. We design three different ways to employ ``dropped residual blocks'' in an SR network and we name them as \texttt{all-part}, \texttt{half-part} and \texttt{quarter-part}. \texttt{all-part} means all the 16 residual blocks are replaced by the ``dropped residual blocks''; \texttt{half-part} means that the second half of the residual blocks are replaced while the others unchanged; and \texttt{quarter} represents only the last four residual blocks are replaced.
\end{enumerate}

\subsection{Dropout Dimension and Probability}
\label{sec:DropoutDimension}
In addition to the position, the dimension of dropout and the probability of dropped channels/elements are also important.
Dropout was originally used for fully-connected layers~\cite{hinton2012improving}; thus there is no need to determine which dimension to drop.
However, after being used in the convolution layers, performing dropout on different dimensions (element and channel) will bring different effects.
We also involve different dropout dimensions in our study.
The element-wise dropout randomly drops elements among all the feature channels, while the channel-wise dropout only randomly drops the entire channels.

Dropout probability determines the percentage of dropped elements or channels.
It is reasonable that too much interference will result in a bad performance, \eg, adding dropout in all blocks or a very high dropout probability.
In a classification network, you can randomly drop up to 50\% of the elements/channels, not affecting the final result but improving generalization performance.
However, this probability may be too large for SR networks as the robustness against information disturbance is much worse than classification networks.
To achieve possible benefits without damaging the network, we first test dropout with probabilities of 10\%, 20\% and 30\%.
We also include higher dropout probabilities (\eg, 50\% or 70\%) in multi-degradation SR.

In total, we have eight optional positions, two dimensions and at least three probabilities to apply dropout in SR networks.
However, most of them are harmful.
Before finally determining our methods, we will study their effects, respectively.
Our results indicate that the \texttt{last-conv} method with channel-wise dropout does not harm SR networks (see Sec.\ref{sec:How to Apply Dropout in SR Networks}).
Therefore, we use this dropout method to exploit the benefits of dropout for multi-degradation SR.

\begin{table*}[ht!]
  \small
  \begin{center}
    \begin{tabular}{|l|c|cc|cc|cc|cc|cc|}
      \hline
      Models                 & Parm. & \multicolumn{2}{c|}{Set5~\cite{Set5}}                                                        & \multicolumn{2}{c|}{Set14~\cite{Set14}}                                                       & \multicolumn{2}{c|}{BSD100~\cite{BSD100}}                                                      & \multicolumn{2}{c|}{Manga109~\cite{Manga109}}                                                    & \multicolumn{2}{c|}{Urban100~\cite{Urban100}}                                                    \\ \hline
                              &            & \multicolumn{1}{c|}{clean}                        & blur                         & \multicolumn{1}{c|}{clean}                        & blur                         & \multicolumn{1}{c|}{clean}                        & blur                         & \multicolumn{1}{c|}{clean}                        & blur                         & \multicolumn{1}{c|}{clean}                        & blur                         \\ \hline
      Real-SRResNet (p=0)    & 1.5M       & \multicolumn{1}{c|}{24.89}                        & 24.76                        & \multicolumn{1}{c|}{23.24}                        & 23.04                        & \multicolumn{1}{c|}{23.89}                        & 23.67                        & \multicolumn{1}{c|}{22.97}                        & 22.59                        & \multicolumn{1}{c|}{21.23}                        & 21.06                        \\ 
      Real-SRResNet (p=0.7) & 1.5M       & \multicolumn{1}{c|}{{\color[HTML]{FF0000} 25.67}} & {\color[HTML]{FF0000} 25.34} & \multicolumn{1}{c|}{{\color[HTML]{FF0000} 23.74}} & {\color[HTML]{FF0000} 23.44} & \multicolumn{1}{c|}{24.18}                        & 23.89                        & \multicolumn{1}{c|}{{\color[HTML]{FF0000} 23.58}} & 22.98                        & \multicolumn{1}{c|}{{\color[HTML]{FF0000} 21.58}} & {\color[HTML]{FF0000} 21.31} \\ 
      Improvement      &            & \multicolumn{1}{c|}{\textbf{+0.78}}               & \textbf{+0.58}               & \multicolumn{1}{c|}{\textbf{+0.50}}               & \textbf{+0.39}               & \multicolumn{1}{c|}{\textbf{+0.29}}               & \textbf{+0.22}               & \multicolumn{1}{c|}{\textbf{+0.61}}               & \textbf{+0.39}               & \multicolumn{1}{c|}{\textbf{+0.35}}               & \textbf{+0.25}               \\ \hline
      Real-RRDB (p=0)         & 16.7M      & \multicolumn{1}{c|}{25.21}                        & 25.14                        & \multicolumn{1}{c|}{23.73}                        & 23.35                        & \multicolumn{1}{c|}{24.42} &  24.22 & \multicolumn{1}{c|}{23.58}                        & {23.16} & \multicolumn{1}{c|}{21.57}                        & 21.17                        \\ 
      Real-RRDB (p=0.5)      & 16.7M      & \multicolumn{1}{c|}{26.05}                        & 26.09                        & \multicolumn{1}{c|}{24.02}                        & 23.96                        & \multicolumn{1}{c|}{24.54}                        & 24.44                        & \multicolumn{1}{c|}{23.78}                        & 23.58                        & \multicolumn{1}{c|}{21.89}                        & 21.75                        \\ 
      Improvement           &            & \multicolumn{1}{c|}{\textbf{+0.84}}               & \textbf{+0.95}               & \multicolumn{1}{c|}{\textbf{+0.29}}               & \textbf{+0.61}               & \multicolumn{1}{c|}{\textbf{+0.12}}               & \textbf{+0.22}               & \multicolumn{1}{c|}{\textbf{+0.20}}               & \textbf{+0.41}               & \multicolumn{1}{c|}{\textbf{+0.32}}               & \textbf{+0.58}               \\ \hline\hline
                              &            & \multicolumn{1}{c|}{noise}                        & jpeg                         & \multicolumn{1}{c|}{noise}                        & jpeg                         & \multicolumn{1}{c|}{noise}                        & jpeg                         & \multicolumn{1}{c|}{noise}                        & jpeg                         & \multicolumn{1}{c|}{noise}                        & jpeg                         \\ \hline
      Real-SRResNet (p=0)    & 1.5M       & \multicolumn{1}{c|}{23.75}                        & 23.70                        & \multicolumn{1}{c|}{22.51}                        & 22.31                        & \multicolumn{1}{c|}{23.01}                        & 23.03                        & \multicolumn{1}{c|}{22.15}                        & 21.75                        & \multicolumn{1}{c|}{20.82}                        & 20.59                        \\ 
      Real-SRResNet (p=0.7) & 1.5M       & \multicolumn{1}{c|}{{\color[HTML]{FF0000} 24.14}} & {\color[HTML]{FF0000} 24.06} & \multicolumn{1}{c|}{22.70}                        & {\color[HTML]{FF0000} 22.64} & \multicolumn{1}{c|}{23.02}                        & 23.24                        & \multicolumn{1}{c|}{{\color[HTML]{FF0000} 22.57}} & 22.03                        & \multicolumn{1}{c|}{20.94}                        & 20.89                        \\ 
      Improvement      &            & \multicolumn{1}{c|}{\textbf{+0.39}}               & \textbf{+0.36}               & \multicolumn{1}{c|}{\textbf{+0.19}}               & \textbf{+0.33}               & \multicolumn{1}{c|}{\textbf{+0.01}}               & \textbf{+0.21}               & \multicolumn{1}{c|}{\textbf{+0.42}}               & \textbf{+0.28}               & \multicolumn{1}{c|}{\textbf{+0.12}}               & \textbf{+0.29}               \\ \hline
      Real-RRDB (p=0)         & 16.7M      & \multicolumn{1}{c|}{24.01}                        & 23.86                        & \multicolumn{1}{c|}{{22.93}} & 22.60                        & \multicolumn{1}{c|}{{23.25}} & {23.33} & \multicolumn{1}{c|}{22.56}                        & {22.18} & \multicolumn{1}{c|}{{21.16}} & {20.92} \\ 
      Real-RRDB (p=0.5)      & 16.7M      & \multicolumn{1}{c|}{24.64}                        & 24.32                        & \multicolumn{1}{c|}{23.17}                        & 22.84                        & \multicolumn{1}{c|}{23.41}                        & 23.42                        & \multicolumn{1}{c|}{22.74}                        & 22.33                        & \multicolumn{1}{c|}{21.26}                        & 21.12                        \\ 
      Improvement           &            & \multicolumn{1}{c|}{\textbf{+0.64}}               & \textbf{+0.46}               & \multicolumn{1}{c|}{\textbf{+0.24}}               & \textbf{+0.24}               & \multicolumn{1}{c|}{\textbf{+0.16}}               & \textbf{+0.10}               & \multicolumn{1}{c|}{\textbf{+0.18}}               & \textbf{+0.16}               & \multicolumn{1}{c|}{\textbf{+0.10}}               & \textbf{+0.19}               \\ \hline\hline
                              &            & \multicolumn{1}{c|}{b+n}                          & b+j                          & \multicolumn{1}{c|}{b+n}                          & b+j                          & \multicolumn{1}{c|}{b+n}                          & b+j                          & \multicolumn{1}{c|}{b+n}                          & b+j                          & \multicolumn{1}{c|}{b+n}                          & b+j                          \\ \hline
      Real-SRResNet (p=0)    & 1.5M       & \multicolumn{1}{c|}{23.20}                        & 23.44                        & \multicolumn{1}{c|}{22.19}                        & 22.06                        & \multicolumn{1}{c|}{22.65}                        & 22.78                        & \multicolumn{1}{c|}{21.56}                        & 21.25                        & \multicolumn{1}{c|}{20.46}                        & 20.29                        \\ 
      Real-SRResNet (p=0.7) & 1.5M       & \multicolumn{1}{c|}{{\color[HTML]{FF0000} 23.47}} & {\color[HTML]{FF0000} 23.69} & \multicolumn{1}{c|}{22.26}                        & {\color[HTML]{FF0000} 22.38} & \multicolumn{1}{c|}{22.60}                        & {\color[HTML]{FF0000} 22.97} & \multicolumn{1}{c|}{{\color[HTML]{FF0000} 21.81}} & 21.45                        & \multicolumn{1}{c|}{20.47}                        & {\color[HTML]{FF0000} 20.53} \\ 
      Improvement      &            & \multicolumn{1}{c|}{\textbf{+0.27}}               & \textbf{+0.25}               & \multicolumn{1}{c|}{\textbf{+0.07}}               & \textbf{+0.32}               & \multicolumn{1}{c|}{\textbf{-0.05}}              & \textbf{+0.19}               & \multicolumn{1}{c|}{\textbf{+0.24}}               & \textbf{+0.20}               & \multicolumn{1}{c|}{\textbf{+0.01}}               & \textbf{+0.23}               \\ \hline
      Real-RRDB (p=0)              & 16.7M      & \multicolumn{1}{c|}{23.40}                        & 23.47                        & \multicolumn{1}{c|}{{22.45}} & 22.17                        & \multicolumn{1}{c|}{{22.77}} & 22.95                        & \multicolumn{1}{c|}{21.74}                        & 21.48 & \multicolumn{1}{c|}{20.57} & 20.39                        \\ 
      Real-RRDB (p=0.5)      & 16.7M      & \multicolumn{1}{c|}{23.73}                        & 23.93                        & \multicolumn{1}{c|}{22.57}                        & 22.59                        & \multicolumn{1}{c|}{22.83}                        & 23.15                        & \multicolumn{1}{c|}{21.76}                        & 21.76                        & \multicolumn{1}{c|}{20.53}                        & 20.69                        \\ 
      Improvement           &            & \multicolumn{1}{c|}{\textbf{+0.33}}               & \textbf{+0.45}               & \multicolumn{1}{c|}{\textbf{+0.12}}               & \textbf{+0.42}               & \multicolumn{1}{c|}{\textbf{+0.06}}               & \textbf{+0.20}               & \multicolumn{1}{c|}{\textbf{+0.02}}               & \textbf{+0.28}               & \multicolumn{1}{c|}{\textbf{-0.04}}               & \textbf{+0.30}               \\ \hline\hline
                              &            & \multicolumn{1}{c|}{n+j}                          & b+n+j                        & \multicolumn{1}{c|}{n+j}                          & b+n+j                        & \multicolumn{1}{c|}{n+j}                          & b+n+j                        & \multicolumn{1}{c|}{n+j}                          & b+n+j                        & \multicolumn{1}{c|}{n+j}                          & b+n+j                        \\ \hline
      Real-SRResNet (p=0)    & 1.5M       & \multicolumn{1}{c|}{23.17}                        & 22.75                        & \multicolumn{1}{c|}{22.01}                        & 21.74                        & \multicolumn{1}{c|}{22.67}                        & 22.39                        & \multicolumn{1}{c|}{21.37}                        & 20.82                        & \multicolumn{1}{c|}{20.41}                        & 20.09                        \\ 
      Real-SRResNet (p=0.7) & 1.5M       & \multicolumn{1}{c|}{ 23.53} & {\color[HTML]{FF0000} 23.04} & \multicolumn{1}{c|}{22.26}                        & {\color[HTML]{FF0000} 21.97} & \multicolumn{1}{c|}{22.81}                        & 22.51                        & \multicolumn{1}{c|}{21.65}                        & 21.03                        & \multicolumn{1}{c|}{20.63}                        & 20.22                        \\ 
      Improvement      &            & \multicolumn{1}{c|}{\textbf{+0.36}}               & \textbf{+0.28}               & \multicolumn{1}{c|}{\textbf{+0.26}}               & \textbf{+0.22}               & \multicolumn{1}{c|}{\textbf{+0.15}}               & \textbf{+0.12}               & \multicolumn{1}{c|}{\textbf{+0.28}}               & \textbf{+0.21}               & \multicolumn{1}{c|}{\textbf{+0.22}}               & \textbf{+0.13}               \\ \hline
      Real-RRDB (p=0)         & 16.7M      & \multicolumn{1}{c|}{23.43}                        & 22.81                        & \multicolumn{1}{c|}{{22.36}} & 21.90                        & \multicolumn{1}{c|}{{22.90}} & {22.51} & \multicolumn{1}{c|}{{21.77}} & {21.05} & \multicolumn{1}{c|}{{20.74}} & {20.23} \\ 
      Real-RRDB (p=0.5)      & 16.7M      & \multicolumn{1}{c|}{23.80}                        & 23.18                        & \multicolumn{1}{c|}{22.49}                        & 22.11                        & \multicolumn{1}{c|}{22.98}                        & 22.61                        & \multicolumn{1}{c|}{21.88}                        & 21.20                        & \multicolumn{1}{c|}{20.83}                        & 20.31                        \\ 
      Improvement           &            & \multicolumn{1}{c|}{\textbf{+0.36}}               & \textbf{+0.37}               & \multicolumn{1}{c|}{\textbf{+0.13}}               & \textbf{+0.20}               & \multicolumn{1}{c|}{\textbf{+0.08}}               & \textbf{+0.10}               & \multicolumn{1}{c|}{\textbf{+0.11}}               & \textbf{+0.15}               & \multicolumn{1}{c|}{\textbf{+0.10}}               & \textbf{+0.08}               \\ \hline
      \end{tabular}
  \end{center}
  \vskip -0.3cm 

  \caption{The PSNR ($\db$) results of models with $\times 4$. Each of two columns gives a test set with 8 types of degradations.
  We apply bicubic, blur, noise and jpeg to generate the degradation, e.g. clean means only bicubic, noise means bicubic $\to$ noise, b+n+j means blur $\to$ bicubic $\to$ noise $\to$ jpeg.
  {\color[HTML]{FF0000}Red} texts mean that the performance of Real-SRResNet (with dropout) is better than Real-RRDB (without dropout), half the test sets are red. $p$ indicates the probability of channel-wise dropout using the \texttt{last-conv} method.}
  
  \label{table:degradations}
  \vskip -0.30cm
  \end{table*}

\section{Experiments}
\label{sec:Experiments}

\subsection{Implementation}
\label{sec:Implementation}
\noindent\textbf{SR Settings.}\quad
There are two commonly-used settings for SR, \ie the single-degradation setting \cite{timofte2017ntire} and the multi-degradation setting \cite{zhang2021designing,wang2021realesrgan}.
The most common degradation used in the single-degradation setting is the bicubic interpolation.
Training and testing under this single-degradation setting can be used to study the capability or performance of the SR networks.
However, SR networks have weak generalization ability under this setting because the network only needs to overfit to a specific degradation.

Unlike the above setting, the multi-degradation setting uses multiple complex degradations to simulate real-world degradations better.
With this setting, the SR networks are expected to be effective in real-world scenarios.
Overfitting to a specific degradation will no longer be suitable in this setting.
The performance of the SR network mainly depends on its generalization ability now.
We follow a successful multi-degradation setting called high-order degradation modelling, which is introduced by Wang \etal \cite{wang2021realesrgan}.
In their setting, complicated combinations of different degradations (\eg,  blurring, downsampling, noising and compression) are used, not one time, but multiple times to generate complex degradations.
All the kernels, downsampling scales, noise and compression, are randomly sampled during the training process on the fly.
We use the same hyperparameters as Wang \etal \cite{wang2021realesrgan,wang2020basicsr}.
As this setting is designed for real-world applications, we use the ``Real'' prefix to represent models trained in this way.

\vspace{2pt}
\noindent\textbf{Training and Testing.}\quad
We use HR images from the DIV2K~\cite{DIV2K} dataset for training.
During training, $L_1$ loss function is adopted with Adam optimizer~\cite{ADAM} ($\beta_1$ = 0.9, $\beta_2$ = 0.999). 
The batch size is 16, LR images are of size 32$\times$32. 
The cosine annealing learning strategy is applied to adjust the learning rate.
The initial learning rate is $2\times10^{-4}$. 
The period of cosine is 500k iterations. 
All models are built using the PyTorch framework~\cite{pytorch} and trained with NVIDIA 2080Ti GPUs.
For testing, we use Set5~\cite{Set5}, Set14~\cite{Set14}, BSD100~\cite{BSD100}, Manga109~\cite{Manga109} and Urban100~\cite{Urban100} as the test sets.
We mainly use PSNR to evaluate the performance of the models \cite{jinjin2020pipal}.
The way to generate LR images in different experiments is also different; we will introduce them in the corresponding sub-sections.

\begin{figure}[ht!]

  \centering
  \subfloat[Channel-wise dropout]{
  \label{fig:decrease12}
  \includegraphics[width=0.94\linewidth{}]{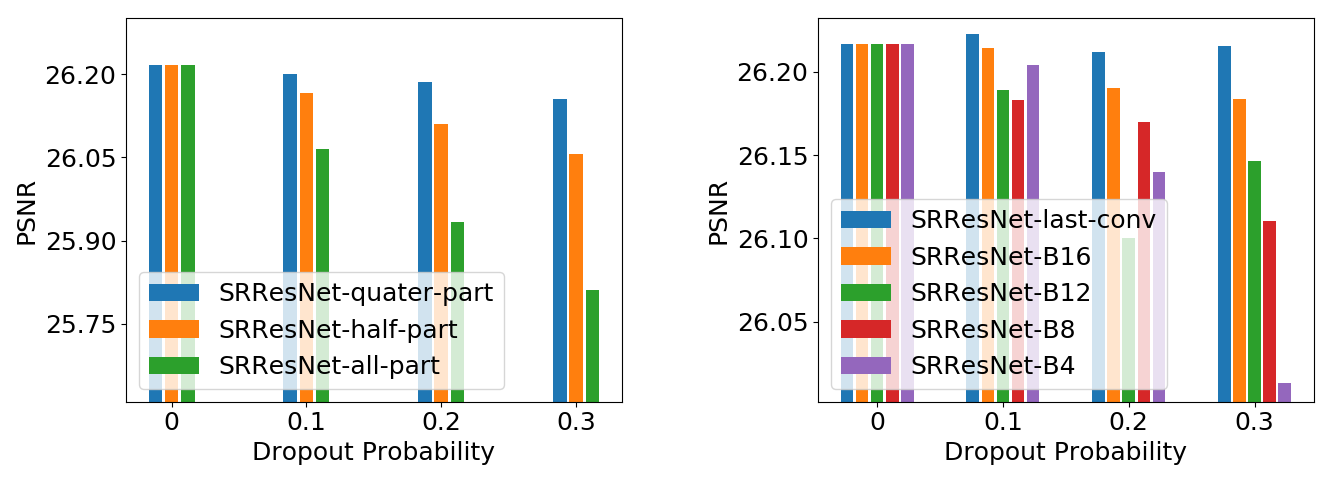}
  }
  
  \subfloat[Element-wise dropout]{
  \label{fig:decrease34}
  \includegraphics[width=0.94\linewidth]{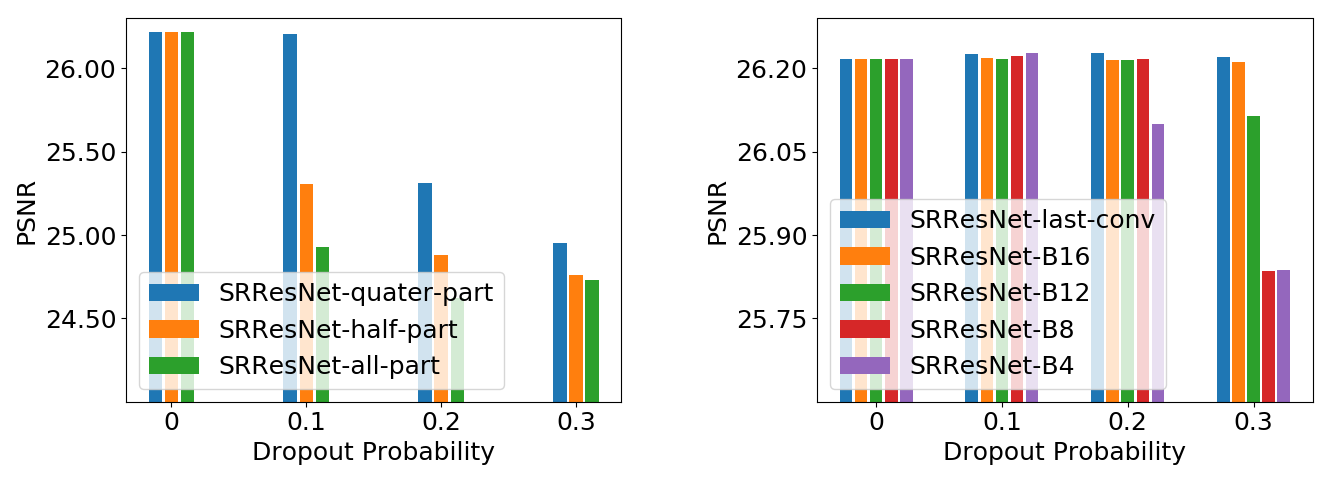}
  }
  \vskip -0.0cm
  \caption{Applying dropout in different methods with various probabilities. The PSNR histograms are obtained by SRResNet on BSD100 with $\times 4$. The training method can be found in Section~\ref{sec:Implementation}. The details of methods can be found in Section~\ref{sec:Dropout}.}
  \label{fig:decrease}
  \vskip -0.2cm
  \end{figure} 
  
\subsection{How to Apply Dropout in SR Networks}
\label{sec:How to Apply Dropout in SR Networks}

We first study the difference between the dropout methods mentioned in Section~\ref{sec:Dropout}.
We test the performance of applying dropout in different ways under the bicubic single-degradation SR setting.
The results are shown in \figurename~\ref{fig:decrease}.
We can obtain the following observations.
Firstly, different dropout positions will lead to completely different performances. 
In the case of using a single dropout layer, we can get better performance when the dropout position comes closer to the output layer. 
When using multiple dropout layers, we can observe larger performance drop for more dropout layers.
Among them, the performance of the \texttt{last-conv} method is the best.
%
%
This observation is consistent with that in classification networks.
Secondly, as can be observed from \figurename~\ref{fig:decrease12} and~\ref{fig:decrease34}, element-wise dropout methods tend to degrade the performance, while channel-wise dropout methods generally perform better.
%
%
Thirdly, in line with expectations, a larger dropout probability will bring worse impact in most cases. 
%
%
%
In conclusion, we propose to apply channel-wise dropout before the last convolution layer.
This position could be easily applied to different network structures, including the vision transformers (included in the supplementary file). 
We find that this simple and straightforward method can already lead to meaningful and robust results, so we adopt this method in the rest of this paper.

\subsection{Dropout in Multi-Degradation SR}
\label{sec:Experiment Results}
Haven the method of applying dropout in SR networks, we next show where we can benefit from it.
Dropout is originally proposed to improve the networks' generalization ability, which perfectly matches our need in developing blind SR networks.
The following experiments demonstrate that dropout does help to train a better blind SR network under the multi-degradation training setting.
In this section, we follow the data generation method proposed by Wang \etal \cite{ESRGAN}, which contains complex degradations and their diverse combinations.

\paragraph{Dropout Helps Learn Better Blind SR Networks.}
\label{sec:Results on Different Degradations}
Under the training setting of multi-degradation, the SR network needs to learn how to restore multiple different degradations simultaneously.
Directly learning to restore all degradations will make the SR networks perform poorly on individual ones.
However, we find that the introduction of dropout can significantly improve the performance of the SR networks under the multi-degradation setting.
We test the performance of dropout in some common degradations and complex degradation combinations.
\tablename~\ref{table:degradations} shows the quantitative comparison of Real-SRResNet and Real-RRDB.
We select Gaussian blur with kernel size 21 and standard deviation 2 (denoted by ``b''), bicubic downsampling, Gaussian noise with a standard deviation 20 (denoted by ``n'') and JPEG compression with quality 50 (denoted by ``j'') as testing degradations.
We also include complex mixed degradations that are combined by the above components.
For these mixed degradations, we synthesize them in the same order as the training method.

\begin{figure}[t]
  \centering
  \includegraphics[width=0.97\linewidth]{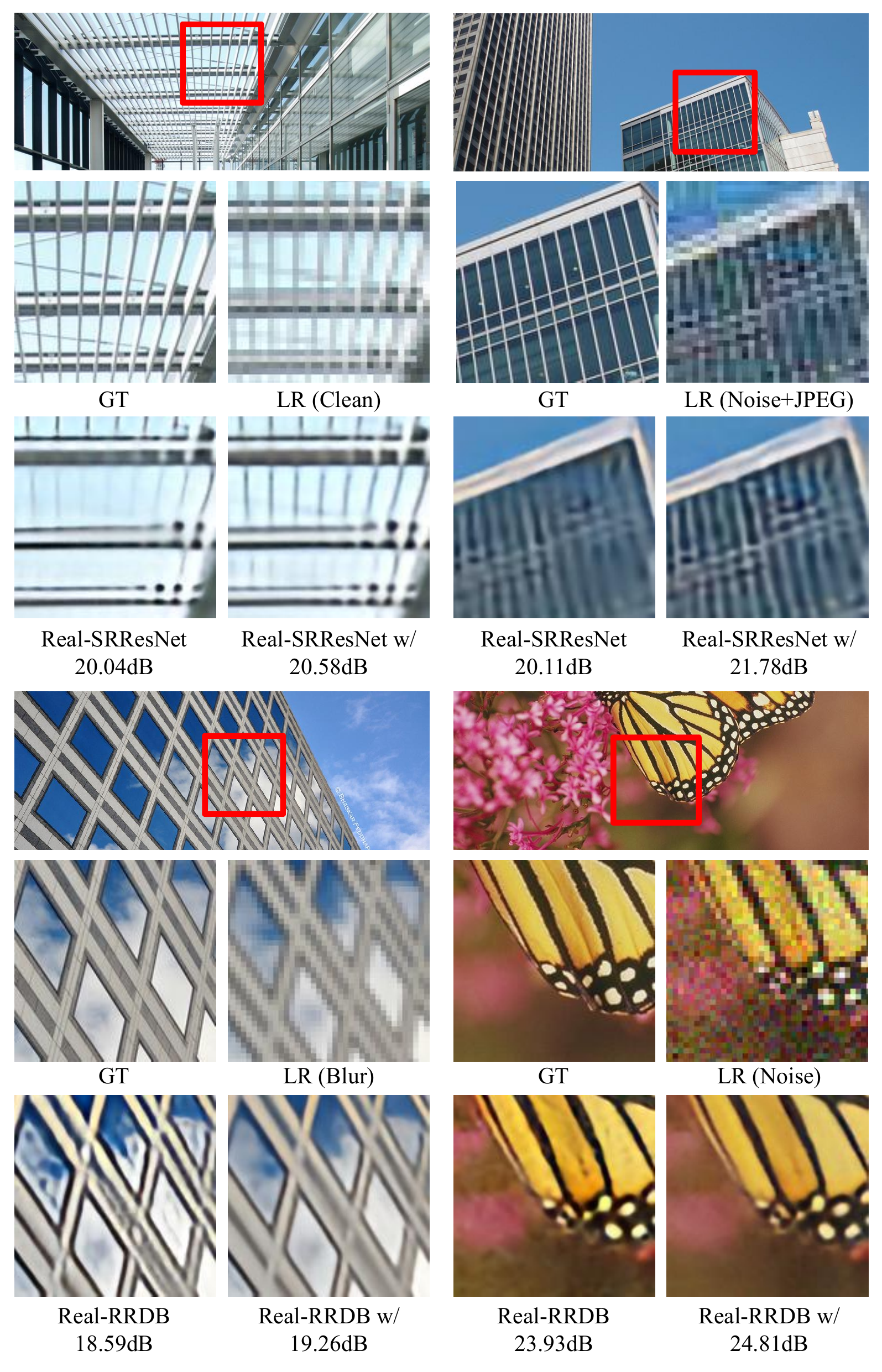}
  \vskip -0.25cm
  \caption{Visual results of representative degradations. We use ``w$\slash$'' to represent the model with dropout in Table~\ref{table:degradations}. (Zoom in for best view)}
  \label{fig:vis}
  \vskip -0.2cm
\end{figure}

When trained with dropout, Real-SRResNet and Real-RRDB obtain better PNSR performance on almost all the datasets with tested degradations.
The maximal improvements on PSNR are 0.78 $\db$  for Real-SRResNet and 0.95 $\db$ for Real-RRDB. 
The red texts mean the performance of Real-SRResNet (with dropout) is better than Real-RRDB. 
An appropriate dropout method makes Real-SRResNet have comparable performance with a much larger model Real-RRDB.
\emph{One line of code is worth a ten-fold increase in the model parameters.}
\figurename~\ref{fig:vis} shows that the models with dropout perform better in content reconstruction, artifact removal and denoising.
The models without dropout may remove or enhance some details incorrectly.

\begin{table}[t!]

  \small %
  \begin{center}

    \begin{tabular}{l|c|c}
\hline
Models    & R-SRResNet (p=0 / 0.7) & R-RRDB (p=0 / 0.5) \\ \hline
mild      & 16.96 / 17.12 (+0.16)    & 16.76 / 17.28 (+0.52)    \\
difficult & 17.81 / 18.01 (+0.20)   &  17.65 / 18.15 (+0.50) \\
wild      & 17.59 / 17.76 (+0.17)   &  17.38 / 17.91 (+0.53) \\ \hline
\end{tabular}

\end{center}
\vskip -0.35cm
\caption{The quantitative comparison (average PSNR) of realistic mild/difficult/wild data in NTIRE 2018 SR challenge\cite {Timofte_2018_CVPR_Workshops}.}
\label{table:downsample}
\vskip -0.5cm
\end{table}

\paragraph{Results on Unseen Degradations.}
\label{sec:Results on Seen and Unseen Degradations}

%
%
%
Theoretically, the test data listed in \tablename~\ref{table:degradations} may be included during training.
To better show the generalization ability improvement after applying the proposed dropout method, we also test the networks using the degradations that are unseen for the networks.
As shown in \tablename~\ref{table:downsample}, dropout also shows superiority when testing on realistic mild/difficult/wild data in NTIRE 2018 SR challenge\cite {Timofte_2018_CVPR_Workshops}. 
It proves that dropout could improve the performance on realistic and unseen degradations.

\paragraph{We show more results} in the supplementary material, including more results of applying dropout with different probabilities and positions under the multi-degradation setting, more visual effect and the performance of applying dropout in a transformer network called SwinIR~\cite{liang2021swinir}.

\section{Interpretation}
\label{sec:Exploration}
After getting the above interesting results, we are very curious about what happens after applying dropout and how dropout improves the network generalization ability.
Next, we investigate the dropout method through the lens of network interpretation and visualization.

\subsection{Dropout Helps Prevent Co-adapting}
\label{sec:Channel Saliency Map}

Dropout is designed to relieve the overfitting problem by preventing co-adapting in high-level vision tasks~\cite{hinton2012improving}.
Many tasks have benefited from using dropout.
Does co-adapting exist in SR tasks? 
Are some features more important for reconstruction than others?
In this section, we investigate these problems, and find that dropout can help SR networks to prevent co-adapting.
The first auxiliary tool we introduce is the channel saliency map (CSM).

\begin{figure}[t!]
  \centering
  \includegraphics[width=0.95\linewidth]{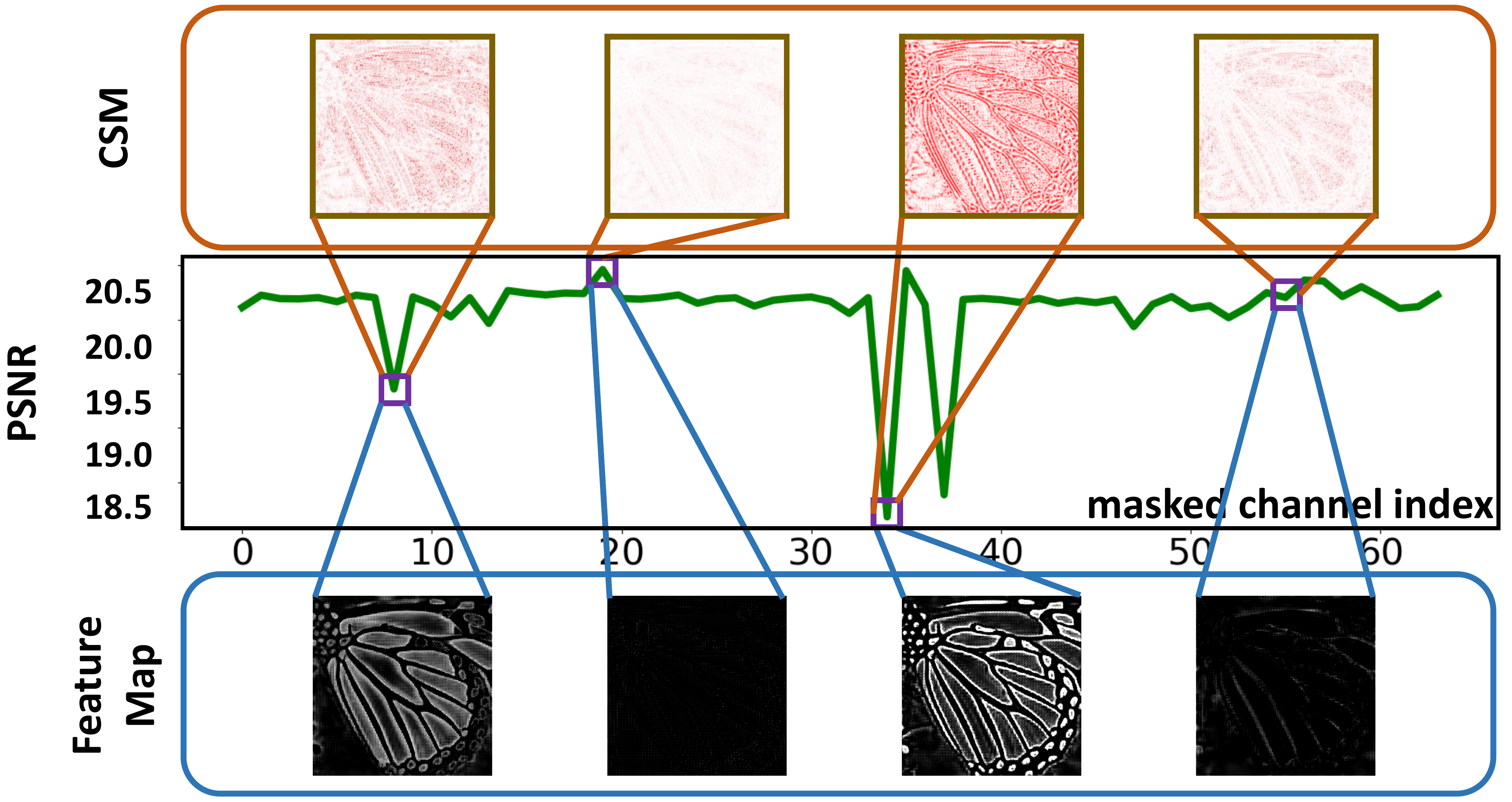}
  \vskip -0.1cm
  \caption{The relationship of PSNR changes, CSM and feature maps. The PSNR of SRResNet without dropout decreases in varying degrees with the ablation of individual channels.}
  \label{fig:attribution}
  \vskip -0.1cm
\end{figure}

Saliency methods \cite{simonyan2013deep, springenberg2014striving, sundararajan2017axiomatic, shrikumar2017learning, lundberg2017unified, gu2021interpreting} are widely used in network interpretation research, which aim at highlighting the important decisive factors of the final output. 
We want to use our CSM method to study different channels' contributions to the final result. 
It is very similar to the previous saliency methods, but we focus on the feature channels.
For an input image $I$, let $F: \mathbb{R}^{h \times w} \rightarrow \mathbb{R}^{s h \times s w}$ be an SR network with the SR factor $s$, ${F}(I)$ be the model output and ${F}_m(I)$ be the intermediate features at layer $m$.
Similar to LAM \cite{gu2021interpreting}, a recent work of localizing important pixels to the SR network output, our goal is to find important feature channels.
One common method to implement attribution analysis is to calculate the gradient of the output value.
Here, we use the summation of image gradient as the attribution target, denoted as $D(I)=\sum{\nabla I}$.
The gradient $\frac{\partial D(I)}{\partial{F}_m(I)}$ reflects the changes of $D(I)$ caused by each element in ${F}_m(I)$, denoted as $Grad_{{F}_m}(I)$.
The higher the gradient is, the more influential the element is.
Note that $Grad_{{F}_m}(I)$ has the same size as ${F}_m(I)$ and also consists of multiple channels.
We remove the sign in $Grad_{{F}_m}(I)$ through an absolute value operation and normalize all its elements to $[0,1]$, as we only need the relative magnitude instead of the real values. 
%
%
We visualize each channel in $Grad_{{F}_m}(I)$ to obtain channel saliency maps.
\figurename~\ref{fig:attribution} shows the relationship between PSNR decrease and saliency maps. 
When we mask different feature maps, we can get different saliency maps and PSNR values.
Low PSNR value is corresponding to bright saliency map.
In the visualization results, a brighter pixel (larger intensity) indicates a larger influence \wrt the SR results. 
It shows some features are significantly more important than others.

\begin{figure}[t!]
  \centering
  \includegraphics[width=0.95\linewidth]{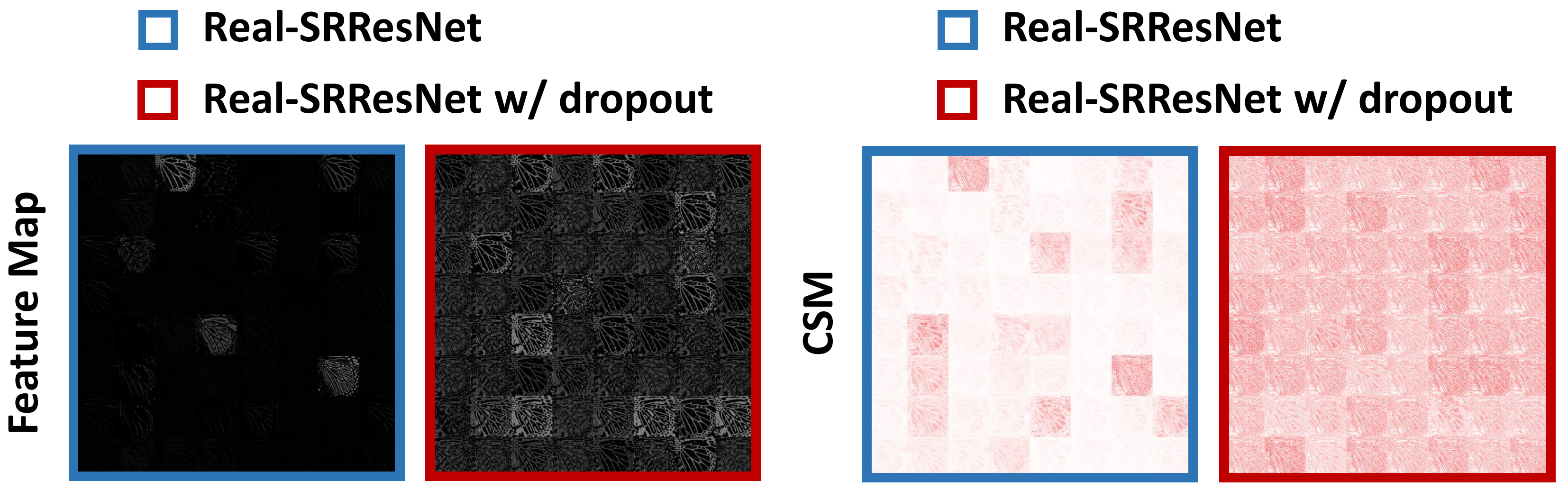}
  \vskip -0.1cm
  \caption{The comparison of feature maps and CSM between Real-SRResNet without dropout and Real-SRResNet with dropout. The features are from the layer where we add dropout.}
  \label{fig:att_feature}
  \vskip -0.1cm
\end{figure}

\begin{figure}[t!]
  \centering
  \includegraphics[width=0.65\linewidth]{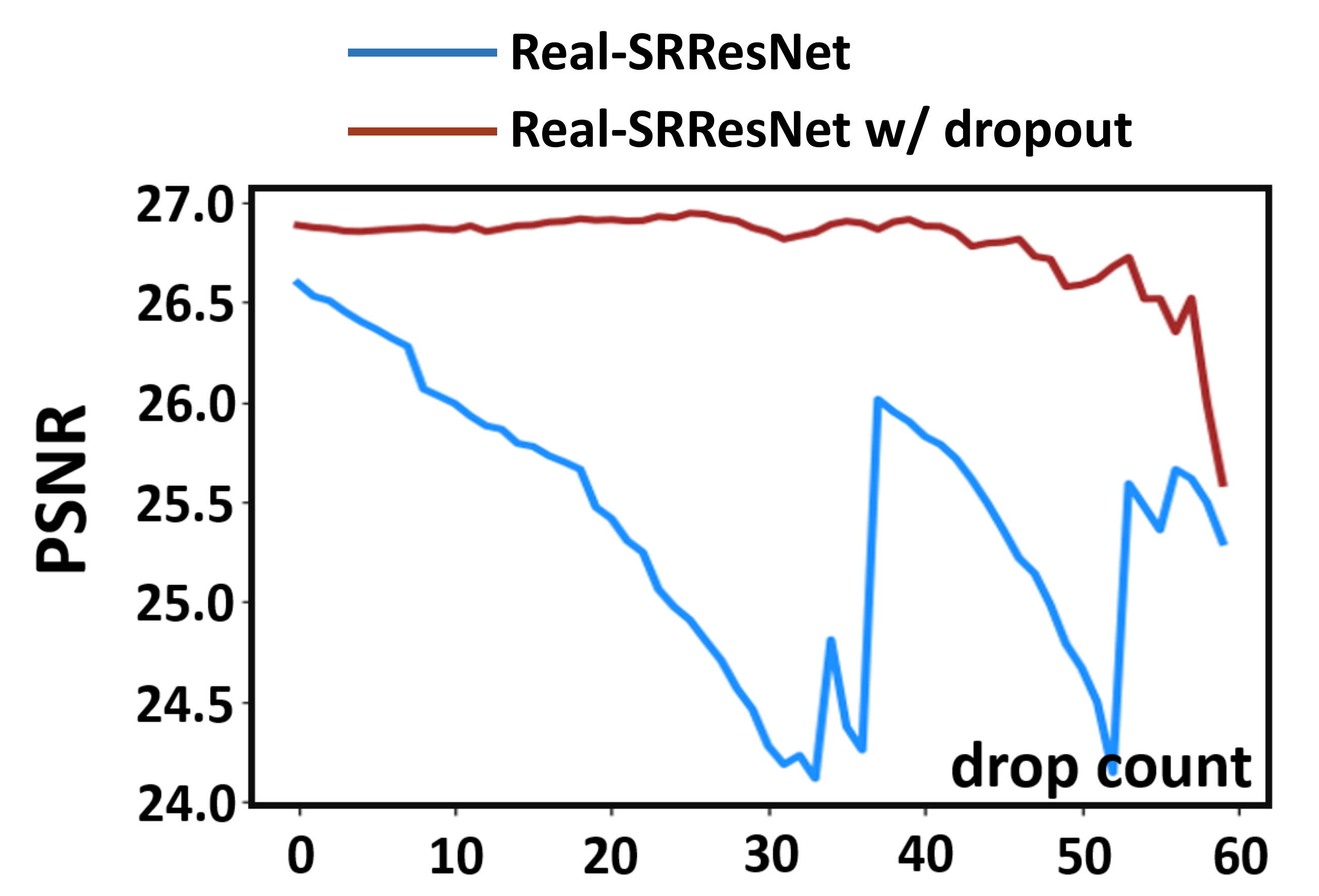}
  \vskip -0.1cm
  \caption{This figure shows PSNR results of channel ablated in turn (from zeroing out one channel to zeroing out 64 channels).}
  \label{fig:line}
  \vskip -0.2cm
\end{figure}

\begin{figure}[t!]
  \centering
  \includegraphics[width=0.99\linewidth]{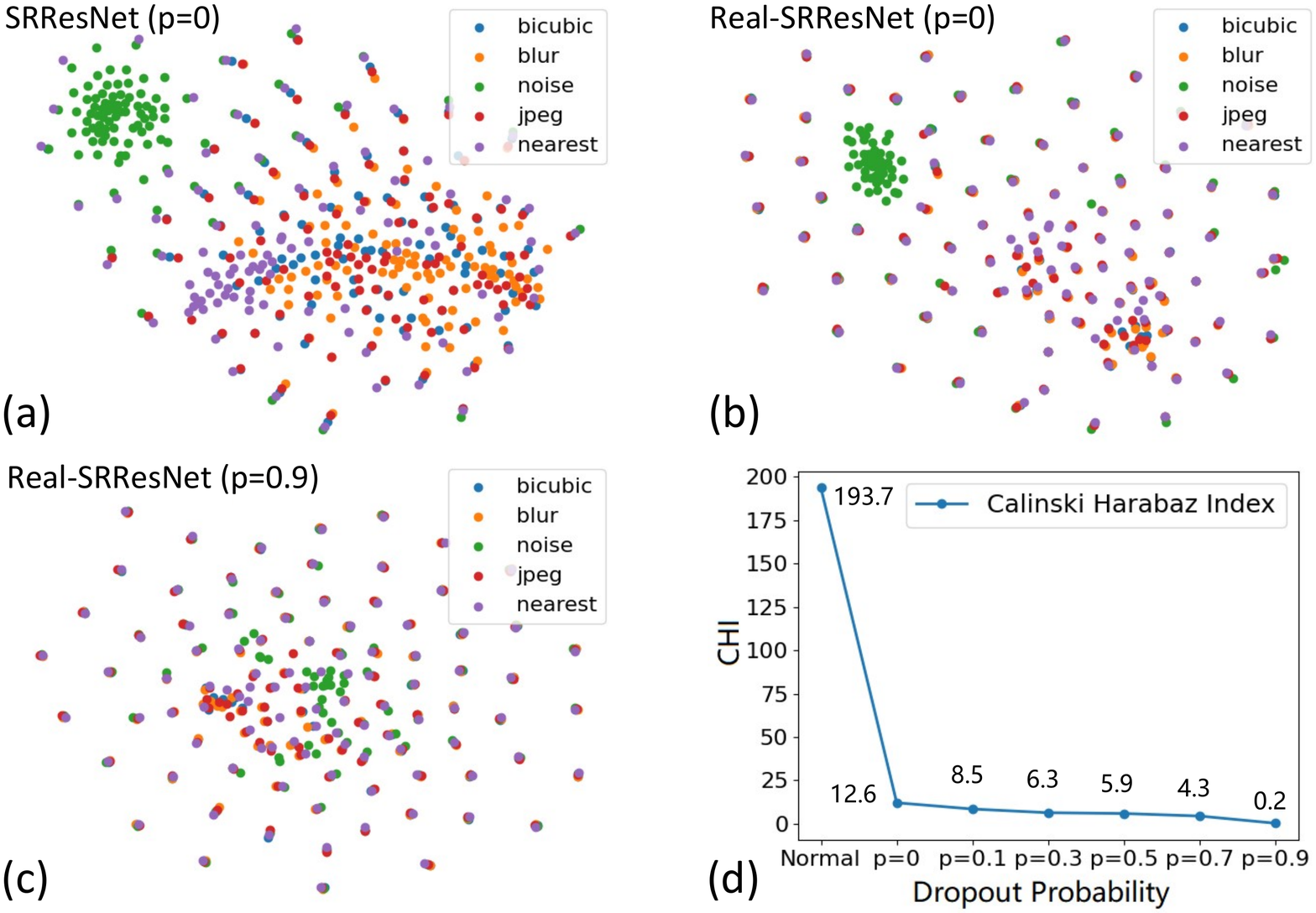} 
 \vskip -0.2cm
  \caption{The DDR clusters of SRResNet and Real-SRResNet with different dropout probabilities. $p=0$ means the networks without dropout. The last subfigure is CHI. $Normal$: SRResNet ($p=0$). With the dropout probability increases, the cluster distributions of different degraded data are more unanimous.}
  \label{fig:DDR_fig}
  \vskip -0.5cm
\end{figure}

A commonly used method called channel ablation~\cite{morcos2018importance} (or filters ablation~\cite{xie2021finding}) also speaks to the same thing.
In practice, we directly ablate an entire feature channel and see what would happen.
We can obtain the importance of each channel by measuring the performance drop once the channel is ablated.
For intermediate features ${F}_m(I)$ with $c$ channels, we have $c$ different choices to zero out an entire channel and then get $c$ ablated results. We use ${F}'_m(I)$ to indicate a ablated result. 
To ensure that the total energy of this layer remains unchanged after ablation, 
each ${F}'_m(I)$ is normalized with $\frac{Sum({F}_m(I))}{Sum({F}'_m(I))}$, where $Sum()$ means summing up all pixel values.
The amplified intermediate features will continue to participate in forwarding calculation until the final output is obtained.
The sharp decrease of PSNR means that the ablated channel contributes more to the output image.
A more important channel will correspond to a brighter feature map in \figurename~\ref{fig:attribution}, this correspondence is coincide with conclusions we have obtained with CSM, that is, some features are more important than others.
Besides, recent works~\cite{xie2021finding,wang2021exploring} also point some features (filters) are more important.

Then we will show that dropout could prevent co-adapting. In other words, dropout could equalize the importance.
%
First, we visualize the feature maps and CSM comparison in \figurename~\ref{fig:att_feature}.
The feature maps and CSM are equalized after adding dropout, it illustrates that dropout could equalize the importance of features.
%
%
To further prove that, we also zero out each channel in turn and linearly scale the rest features with $\frac{Sum({F}_m(I))}{Sum({F}'_m(I))}$. 
\figurename~\ref{fig:line} shows that the PSNR values of Real-SRResNet without dropout would decrease severely with more channels being ablated, but the performance of Real-SRResNet with dropout keeps unchanged. 
For a model with dropout, PSNR no longer depends on several specific channels. Even one-third channels of the network are enough to maintain performance.
It also show that dropout could equalize the channel importance.

The above experiments demonstrate that dropout can help SR networks to prevent co-adapting.

\subsection{Dropout Helps Improve Generalization Ability}
\label{sec:Dropout Helps Improve Generalization Ability}
The most direct strategy to evaluate generalization ability is to test models in a wide range of data, as described in Section~\ref{sec:Results on Different Degradations}.
It is hard to predict the model's generalization performance for images and degradations that have not been tested -- maybe the model happens to perform well on the tested data.
However, there are also methods to evaluate generalization ability from the view of interpreting networks' behaviours.

In low-level vision, Liu \etal~\cite{liu2021discovering} present a concept called deep degradation representation (DDR).
Here, we will refer to \figurename~\ref{fig:DDR_fig} when introducing DDR. Each point in \figurename~\ref{fig:DDR_fig}{\color[HTML]{FF0000}a},~\ref{fig:DDR_fig}{\color[HTML]{FF0000}b} and \ref{fig:DDR_fig}{\color[HTML]{FF0000}c} represents an input sample ($128\times128$ image). 
There are 500 points in each sub-figure.
These samples are produced from five degradations, and each degradation corresponding to the same 100 images.
DDR reveals that SR networks could classify the inputs to different ``degradation semantics''.
For example, in \figurename~\ref{fig:DDR_fig}{\color[HTML]{FF0000}a}, points with different colors indicate the inputs with different degradations. 
Inputs with same degradations (points with same colors) will be clustered.
If the obtained clusters are well divided, the network tends to only process specific degradation clusters and ignore other clusters, resulting in poor generalization performance.
If the clustering trend is weak, the network has handled all the inputs well.
For example, as can be observed from the comparison of \figurename~\ref{fig:DDR_fig}{\color[HTML]{FF0000}a} and \figurename~\ref{fig:DDR_fig}{\color[HTML]{FF0000}b}, the clustering degree of the original SRResNet without dropout is larger than Real-SRResNet.
This illustrates that a network that has seen more degradations has more remarkable generalization ability.

When it comes to dropout, the cluster distributions of different degraded data for Real-SRResNet ($p=0.9$, \figurename~\ref{fig:DDR_fig}{\color[HTML]{FF0000}c}) are closer than Real-SRResNet ($p=0.1$, \figurename~\ref{fig:DDR_fig}{\color[HTML]{FF0000}b}).
Besides directly observing distribution maps, we could also use Calinski-Harabaz Index (CHI)~\cite{calinski1974dendrite} to measure the separation degree of clusters. 
Lower CHI means weaker clustering degree, which also indicates better generalization ability.
In \figurename~\ref{fig:DDR_fig}{\color[HTML]{FF0000}d}, one can observe that CHI decreases with the dropout probability increases.
It demonstrates that dropout improves the generalization ability of the SR network.
This phenomenon is a mutual corroboration with our testing results in a wide range of data.
Another interesting observation is that the distribution of samples with noise (the green points in \figurename~\ref{fig:DDR_fig}) is always the most different one.
Reflected in the restoration performance mentioned in Section~\ref{sec:Results on Different Degradations}, the performance obtained on noisy data is also far from that on clean.


\section{Conclusion}
\label{sec:Conclusion}

In this work, we explore the usage and working mechanism of dropout in SR task.
Specifically, we discover that adding dropout using \texttt{last-conv} method can significantly improve the network performance in the multi-degradation setting.
As for the working mechanism, we find that dropout indeed improves the representation ability of channels and the generalization ability of networks.
This is a mutual corroboration of our experimental results.
We believe that this work will bring a new perspective to SR tasks and help us better understand network behaviours.
%

\paragraph{Acknowledgements.} This work is partially supported by the National Natural Science Foundation of China (61906184), the Joint Lab of CAS-HK, the Shenzhen Research Program (RCJC20200714114557087), the Shanghai Committee of Science and Technology, China (Grant No. 21DZ1100100).

{\small
\bibliographystyle{ieee_fullname}
\bibliography{egbib}
}

\clearpage
\renewcommand\thesection{\Alph{section}}
	\renewcommand\thesubsection{\thesection.\arabic{subsection}}
	\renewcommand\thefigure{\Alph{section}.\arabic{figure}}
	\renewcommand\thetable{\Alph{section}.\arabic{table}} 
	
	\noindent{\large{\textbf{Supplementary Materials}}}
	
	\setcounter{section}{0}
	\setcounter{figure}{0}
	\setcounter{table}{0}

In this supplementary file, we first apply the proposed dropout method to SwinIR~\cite{liang2021swinir}, which is a transformer-based SR backbone network. The experimental results show dropout is also helpful for transformer-based SR networks. Second, we provide more results of using different dropout probabilities and dropout positions under multi-degradation setting. Then, we show some training curves to illustrate that dropout does not change the convergence trend. Finally, we show more qualitative results to show the effectiveness of dropout.

    \section{Applying Dropout in SwinIR}
    \label{sec:Applying Dropout in SwinIR}
    \begin{table*}[ht!]
      \small
      \begin{center}
          \begin{tabular}{|l|cc|cc|cc|cc|cc|}
              \hline
              Models              & \multicolumn{2}{c|}{Set5~\cite{Set5}}                          & \multicolumn{2}{c|}{Set14~\cite{Set14}}                         & \multicolumn{2}{c|}{BSD100~\cite{BSD100}}                          & \multicolumn{2}{c|}{Manga109~\cite{Manga109}}                        & \multicolumn{2}{c|}{Urban100~\cite{Urban100}}                      \\ \hline
                                  & \multicolumn{1}{c|}{clean}         & blur          & \multicolumn{1}{c|}{clean}         & blur          & \multicolumn{1}{c|}{clean}          & blur           & \multicolumn{1}{c|}{clean}          & blur           & \multicolumn{1}{c|}{clean}         & blur          \\ \hline
              Real-SwinIR (p=0)   & \multicolumn{1}{c|}{25.58}         & 25.50         & \multicolumn{1}{c|}{23.89}         & 23.68         & \multicolumn{1}{c|}{24.43}          & 24.23          & \multicolumn{1}{c|}{23.80}          & 23.53          & \multicolumn{1}{c|}{21.73}         & 21.57         \\ 
              Real-SwinIR (p=0.5) & \multicolumn{1}{c|}{26.04}         & 25.78         & \multicolumn{1}{c|}{23.97}         & 23.69         & \multicolumn{1}{c|}{24.44}          & 24.19          & \multicolumn{1}{c|}{23.88}          & 23.55          & \multicolumn{1}{c|}{21.86}         & 21.67         \\ \hline
              Improvement         & \multicolumn{1}{c|}{\textbf{+0.46}} & \textbf{+0.29} & \multicolumn{1}{c|}{\textbf{+0.08}} & \textbf{+0.01} & \multicolumn{1}{c|}{\textbf{+0.01}}  & \textbf{-0.04} & \multicolumn{1}{c|}{\textbf{+0.08}}  & \textbf{+0.03}  & \multicolumn{1}{c|}{\textbf{+0.12}} & \textbf{+0.10} \\ \hline\hline
                                  & \multicolumn{1}{c|}{noise}         & jepg          & \multicolumn{1}{c|}{noise}         & jepg          & \multicolumn{1}{c|}{noise}          & jepg           & \multicolumn{1}{c|}{noise}          & jepg           & \multicolumn{1}{c|}{noise}         & jepg          \\ \hline
              Real-SwinIR (p=0)   & \multicolumn{1}{c|}{24.40}         & 24.03         & \multicolumn{1}{c|}{22.97}         & 22.71         & \multicolumn{1}{c|}{23.40}          & 23.34          & \multicolumn{1}{c|}{22.83}          & 22.27          & \multicolumn{1}{c|}{21.20}         & 20.95         \\ 
              Real-SwinIR (p=0.5) & \multicolumn{1}{c|}{24.64}         & 24.32         & \multicolumn{1}{c|}{23.10}         & 22.86         & \multicolumn{1}{c|}{23.42}          & 23.40          & \multicolumn{1}{c|}{22.79}          & 22.34          & \multicolumn{1}{c|}{21.35}         & 21.11         \\ \hline
              Improvement         & \multicolumn{1}{c|}{\textbf{+0.24}} & \textbf{+0.30} & \multicolumn{1}{c|}{\textbf{+0.13}} & \textbf{+0.15} & \multicolumn{1}{c|}{\textbf{+0.03}}  & \textbf{+0.06}  & \multicolumn{1}{c|}{\textbf{-0.03}} & \textbf{+0.07}  & \multicolumn{1}{c|}{\textbf{+0.15}} & \textbf{+0.16} \\ \hline\hline
                                  & \multicolumn{1}{c|}{b+n}           & b+j           & \multicolumn{1}{c|}{b+n}           & b+j           & \multicolumn{1}{c|}{b+n}            & b+j            & \multicolumn{1}{c|}{b+n}            & b+j            & \multicolumn{1}{c|}{b+n}           & b+j           \\ \hline
              Real-SwinIR (p=0)   & \multicolumn{1}{c|}{23.64}         & 23.67         & \multicolumn{1}{c|}{22.48}         & 22.43         & \multicolumn{1}{c|}{22.94}          & 23.08          & \multicolumn{1}{c|}{22.11}          & 21.72          & \multicolumn{1}{c|}{20.71}         & 20.59         \\ 
              Real-SwinIR (p=0.5) & \multicolumn{1}{c|}{23.80}         & 23.84         & \multicolumn{1}{c|}{22.59}         & 22.54         & \multicolumn{1}{c|}{22.89}          & 23.10          & \multicolumn{1}{c|}{22.01}          & 21.77          & \multicolumn{1}{c|}{20.77}         & 20.71         \\ \hline
              Improvement         & \multicolumn{1}{c|}{\textbf{+0.17}} & \textbf{+0.17} & \multicolumn{1}{c|}{\textbf{+0.11}} & \textbf{+0.11} & \multicolumn{1}{c|}{\textbf{-0.05}} & \textbf{+0.02}  & \multicolumn{1}{c|}{\textbf{-0.10}} & \textbf{+0.04}  & \multicolumn{1}{c|}{\textbf{+0.06}} & \textbf{+0.12} \\ \hline\hline
                                  & \multicolumn{1}{c|}{n+j}           & b+n+j         & \multicolumn{1}{c|}{n+j}           & b+n+j         & \multicolumn{1}{c|}{n+j}            & b+n+j          & \multicolumn{1}{c|}{n+j}            & b+n+j          & \multicolumn{1}{c|}{n+j}           & b+n+j         \\ \hline
              Real-SwinIR (p=0)   & \multicolumn{1}{c|}{23.45}         & 22.91         & \multicolumn{1}{c|}{22.29}         & 21.96         & \multicolumn{1}{c|}{22.86}          & 22.53          & \multicolumn{1}{c|}{21.80}          & 21.17          & \multicolumn{1}{c|}{20.67}         & 20.28         \\ 
              Real-SwinIR (p=0.5) & \multicolumn{1}{c|}{23.67}         & 23.10         & \multicolumn{1}{c|}{22.44}         & 22.08         & \multicolumn{1}{c|}{22.89}          & 22.51          & \multicolumn{1}{c|}{21.73}          & 21.11          & \multicolumn{1}{c|}{20.81}         & 20.35         \\ \hline
              Improvement         & \multicolumn{1}{c|}{\textbf{+0.22}} & \textbf{+0.19} & \multicolumn{1}{c|}{\textbf{+0.14}} & \textbf{+0.12} & \multicolumn{1}{c|}{\textbf{+0.03}}  & \textbf{-0.02} & \multicolumn{1}{c|}{\textbf{-0.07}} & \textbf{-0.06} & \multicolumn{1}{c|}{\textbf{+0.14}} & \textbf{+0.07} \\ \hline
              \end{tabular}
      \end{center}
      \vskip -0.3cm
      \caption{The PSNR ($\db$) results of Real-SwinIR with $\times 4$. Each of two columns gives a test set with 8 types of degradations. We apply bicubic, blur, noise and jpeg to generate the degradation, e.g. clean means only bicubic, noise means bicubic $\to$ noise, b+n+j means blur $\to$ bicubic $\to$ noise $\to$ jpeg.}
      \vskip -0.3cm
      \label{table:SwinIR}
      \end{table*}
    
SwinIR~\cite{liang2021swinir} is a newly proposed SR backbone network using the transformer mechanism. This model achieves state-of-the-art performance in many restoration tasks. We also apply the dropout method to this model to demonstrate that dropout is also helpful for transformer-based SR models.
    
We apply the dropout layer before the output convolutional layer (from 64 channels to 3 channels, \texttt{last-conv}). SwinIR also has this structure. We use the same training and testing data as Real-SRResNet and Real-RRDB for Real-SwinIR. The original setting of SwinIR that the $\times 4$ model is finetuned from the $\times 2$ model needs a too long training time. Therefore, we follow the reproduction~\cite{wang2020basicsr} to train the models from scratch and also show the results of 250K iteration just like this reproduction. Note that, we only train the model with dropout ($p=0.5$) to make a simple verification. This training setting and dropout probability may not be the most appropriate for SwinIR but are enough to illustrate dropout is also helpful.

The results are shown in \tablename~\ref{table:SwinIR}. When trained with dropout, Real-SwinIR obtains better PNSR performance on most of the five datasets with the tested degradations. The maximal improvement on PSNR is 0.46 $\db$.

\begin{table}[t!]

  \small %
  \begin{center}

    \begin{tabular}{|c|c|c|c|c|c|c|}
      \hline
      Prob.  & p=0   & p=0.1 & p=0.3 & p=0.5 & p=0.7 & p=0.9 \\ \hline\hline
      Set1 & 22.15 & 22.31 & 22.35 & {\color[HTML]{0000FF}22.51} & {\color[HTML]{FF0000}22.57} & 22.31 \\ \hline
      Set2 & 20.82 & 20.85 & 20.88 & {\color[HTML]{FF0000}20.97} & {\color[HTML]{0000FF}20.94} & 20.64 \\ \hline
      \end{tabular}
  
\end{center}
\vskip -0.45cm
\caption{The performance of using different dropout probabilities for Real-SRResNet with $\times 4$.  
Set1 is Manga109 with noise and Set2 is Urban100 with noise (standard deviation is 20). {\color[HTML]{FF0000}Red}/{\color[HTML]{0000FF}Blue} text: best/second-best PSNR ($\db$).}
\label{table:probabilities}
\vskip -0.35cm
\end{table}

\section{Ablation Study on Dropout Positions and Probabilities }
\label{sec:Results on Different Dropout Probabilities}
We propose to apply channel-wise dropout before the last convolution layer under multi-degradation in main experiments. Beside, we also provide experiments on different dropout positions and probabilities under multi-degradation.

\paragraph{Positions.}
We show the performance of Real-SRResNet with different dropout positions in \tablename~\ref{table:positions}. 
Most dropout methods can improve the performance except \texttt{half-part} and \texttt{all-part} methods. The \texttt{last-conv} method obtains most of the best results (red text).
So we chose \texttt{last-conv} method in main paper. This position is a safe and general choice, which can maintain the network capacity and improve the generalization ability. 
Besides, this position can be easily applied to different network structures, including Transformer, while Dropblock cannot. 
This simple and straightforward method can already lead to meaningful and robust results.

\begin{table*}[ht!]
      \small
      \begin{center}
      \begin{tabular}{|l|cc|cc|cc|cc|cc|}
\hline
\multicolumn{1}{|c|}{} & \multicolumn{2}{c|}{Set5}                                                        & \multicolumn{2}{c|}{Set14}                                                       & \multicolumn{2}{c|}{BSD100}                                                      & \multicolumn{2}{c|}{Manga109}                                                    & \multicolumn{2}{c|}{Urban100}                                                    \\ \hline
\multicolumn{1}{|c|}{} & \multicolumn{1}{c|}{clean}                        & blur                         & \multicolumn{1}{c|}{clean}                        & blur                         & \multicolumn{1}{c|}{clean}                        & blur                         & \multicolumn{1}{c|}{clean}                        & blur                         & \multicolumn{1}{c|}{clean}                        & blur                         \\ \hline
Real-SRResNet (p=0)    & \multicolumn{1}{c|}{24.89}                        & 24.76                        & \multicolumn{1}{c|}{23.24}                        & 23.04                        & \multicolumn{1}{c|}{23.89}                        & 23.67                        & \multicolumn{1}{c|}{22.97}                        & 22.59                        & \multicolumn{1}{c|}{21.23}                        & 21.06                        \\
last (p=0.7)           & \multicolumn{1}{c|}{25.67}                        & 25.34                        & \multicolumn{1}{c|}{23.74}                        & 23.44                        & \multicolumn{1}{c|}{24.18}                        & 23.89                        & \multicolumn{1}{c|}{23.58}                        & {\color[HTML]{FF0000} 22.98} & \multicolumn{1}{c|}{21.58}                        & {\color[HTML]{FF0000} 21.31} \\
B4 (p=0.7)             & \multicolumn{1}{c|}{25.68}                        & 24.93                        & \multicolumn{1}{c|}{23.95}                        & 23.40                        & \multicolumn{1}{c|}{24.25}                        & 23.74                        & \multicolumn{1}{c|}{23.49}                        & {\color[HTML]{1D41D5} 22.61} & \multicolumn{1}{c|}{21.66}                        & 21.08                        \\
B8 (p=0.7)             & \multicolumn{1}{c|}{26.59}                        & 25.42                        & \multicolumn{1}{c|}{24.38}                        & 23.62                        & \multicolumn{1}{c|}{24.61}                        & 23.95                        & \multicolumn{1}{c|}{{\color[HTML]{1D41D5} 23.80}} & 22.54                        & \multicolumn{1}{c|}{21.98}                        & 21.24                        \\
B12 (p=0.7)            & \multicolumn{1}{c|}{{\color[HTML]{FF0000} 26.78}} & {\color[HTML]{FF0000} 25.59} & \multicolumn{1}{c|}{{\color[HTML]{1D41D5} 24.44}} & {\color[HTML]{FF0000} 23.70} & \multicolumn{1}{c|}{{\color[HTML]{FF0000} 24.64}} & {\color[HTML]{FF0000} 24.00} & \multicolumn{1}{c|}{{\color[HTML]{FF0000} 23.81}} & 22.54                        & \multicolumn{1}{c|}{{\color[HTML]{FF0000} 22.04}} & {\color[HTML]{1D41D5} 21.28} \\
B16 (p=0.7)            & \multicolumn{1}{c|}{{\color[HTML]{1D41D5} 26.75}} & {\color[HTML]{1D41D5} 25.48} & \multicolumn{1}{c|}{{\color[HTML]{FF0000} 24.48}} & {\color[HTML]{1D41D5} 23.66} & \multicolumn{1}{c|}{{\color[HTML]{1D41D5} 24.64}} & {\color[HTML]{1D41D5} 23.97} & \multicolumn{1}{c|}{23.77}                        & 22.49                        & \multicolumn{1}{c|}{{\color[HTML]{1D41D5} 22.01}} & 21.20                        \\
quarter (p=0.7)        & \multicolumn{1}{c|}{25.10}                        & 24.66                        & \multicolumn{1}{c|}{23.22}                        & 22.87                        & \multicolumn{1}{c|}{23.64}                        & 23.32                        & \multicolumn{1}{c|}{22.77}                        & 22.07                        & \multicolumn{1}{c|}{21.04}                        & 20.74                        \\
half (p=0.7)           & \multicolumn{1}{c|}{25.84}                        & 24.25                        & \multicolumn{1}{c|}{23.90}                        & 22.78                        & \multicolumn{1}{c|}{24.20}                        & 23.31                        & \multicolumn{1}{c|}{22.56}                        & 21.18                        & \multicolumn{1}{c|}{21.25}                        & 20.30                        \\
all (p=0.7)            & \multicolumn{1}{c|}{25.82}                        & 24.24                        & \multicolumn{1}{c|}{23.89}                        & 22.77                        & \multicolumn{1}{c|}{24.19}                        & 23.31                        & \multicolumn{1}{c|}{22.56}                        & 21.18                        & \multicolumn{1}{c|}{21.24}                        & 20.30                        \\ \hline
                       & \multicolumn{1}{c|}{noise}                        & jepg                         & \multicolumn{1}{c|}{noise}                        & jepg                         & \multicolumn{1}{c|}{noise}                        & jepg                         & \multicolumn{1}{c|}{noise}                        & jepg                         & \multicolumn{1}{c|}{noise}                        & jepg                         \\ \hline
Real-SRResNet (p=0)    & \multicolumn{1}{c|}{23.75}                        & 23.70                        & \multicolumn{1}{c|}{22.51}                        & 22.31                        & \multicolumn{1}{c|}{{\color[HTML]{1D41D5} 23.01}} & 23.03                        & \multicolumn{1}{c|}{22.15}                        & 21.75                        & \multicolumn{1}{c|}{20.82}                        & 20.59                        \\
last (p=0.7)           & \multicolumn{1}{c|}{{\color[HTML]{FF0000} 24.14}} & 24.06                        & \multicolumn{1}{c|}{{\color[HTML]{1D41D5} 22.70}} & 22.64                        & \multicolumn{1}{c|}{{\color[HTML]{FF0000} 23.02}} & 23.24                        & \multicolumn{1}{c|}{{\color[HTML]{FF0000} 22.57}} & {\color[HTML]{FF0000} 22.03} & \multicolumn{1}{c|}{{\color[HTML]{FF0000} 20.94}} & 20.89                        \\
B4 (p=0.7)             & \multicolumn{1}{c|}{{\color[HTML]{1D41D5} 24.00}} & 24.05                        & \multicolumn{1}{c|}{{\color[HTML]{FF0000} 22.73}} & 22.84                        & \multicolumn{1}{c|}{22.93}                        & 23.36                        & \multicolumn{1}{c|}{{\color[HTML]{1D41D5} 22.35}} & {\color[HTML]{1D41D5} 21.92} & \multicolumn{1}{c|}{{\color[HTML]{1D41D5} 20.83}} & 20.96                        \\
B8 (p=0.7)             & \multicolumn{1}{c|}{23.73}                        & 24.33                        & \multicolumn{1}{c|}{22.40}                        & 22.96                        & \multicolumn{1}{c|}{22.47}                        & {\color[HTML]{1D41D5} 23.45} & \multicolumn{1}{c|}{21.99}                        & 21.79                        & \multicolumn{1}{c|}{20.60}                        & 21.08                        \\
B12 (p=0.7)            & \multicolumn{1}{c|}{23.91}                        & {\color[HTML]{FF0000} 24.38} & \multicolumn{1}{c|}{22.52}                        & {\color[HTML]{FF0000} 22.98} & \multicolumn{1}{c|}{22.61}                        & {\color[HTML]{FF0000} 23.47} & \multicolumn{1}{c|}{22.11}                        & 21.76                        & \multicolumn{1}{c|}{20.70}                        & {\color[HTML]{FF0000} 21.12} \\
B16 (p=0.7)            & \multicolumn{1}{c|}{23.91}                        & {\color[HTML]{1D41D5} 24.37} & \multicolumn{1}{c|}{22.53}                        & {\color[HTML]{1D41D5} 22.97} & \multicolumn{1}{c|}{22.61}                        & 23.44                        & \multicolumn{1}{c|}{22.06}                        & 21.73                        & \multicolumn{1}{c|}{20.69}                        & {\color[HTML]{1D41D5} 21.08} \\
quarter (p=0.7)        & \multicolumn{1}{c|}{22.97}                        & 23.50                        & \multicolumn{1}{c|}{21.71}                        & 22.15                        & \multicolumn{1}{c|}{21.90}                        & 22.85                        & \multicolumn{1}{c|}{21.47}                        & 21.09                        & \multicolumn{1}{c|}{20.06}                        & 20.43                        \\
half (p=0.7)           & \multicolumn{1}{c|}{22.65}                        & 23.54                        & \multicolumn{1}{c|}{21.56}                        & 22.40                        & \multicolumn{1}{c|}{21.70}                        & 23.02                        & \multicolumn{1}{c|}{20.80}                        & 20.61                        & \multicolumn{1}{c|}{19.77}                        & 20.40                        \\
all (p=0.7)            & \multicolumn{1}{c|}{22.64}                        & 23.53                        & \multicolumn{1}{c|}{21.56}                        & 22.40                        & \multicolumn{1}{c|}{21.69}                        & 23.02                        & \multicolumn{1}{c|}{20.79}                        & 20.60                        & \multicolumn{1}{c|}{19.77}                        & 20.40                        \\ \hline
                       & \multicolumn{1}{c|}{b+n}                          & b+j                          & \multicolumn{1}{c|}{b+n}                          & b+j                          & \multicolumn{1}{c|}{b+n}                          & b+j                          & \multicolumn{1}{c|}{b+n}                          & b+j                          & \multicolumn{1}{c|}{b+n}                          & b+j                          \\ \hline
Real-SRResNet (p=0)    & \multicolumn{1}{c|}{23.20}                        & 23.44                        & \multicolumn{1}{c|}{22.19}                        & 22.06                        & \multicolumn{1}{c|}{{\color[HTML]{FF0000} 22.65}} & 22.78                        & \multicolumn{1}{c|}{{\color[HTML]{1D41D5} 21.56}} & {\color[HTML]{1D41D5} 21.25} & \multicolumn{1}{c|}{{\color[HTML]{1D41D5} 20.46}} & 20.29                        \\
last (p=0.7)           & \multicolumn{1}{c|}{{\color[HTML]{FF0000} 23.47}} & {\color[HTML]{FF0000} 23.69} & \multicolumn{1}{c|}{{\color[HTML]{FF0000} 22.26}} & {\color[HTML]{3531FF} 22.38} & \multicolumn{1}{c|}{{\color[HTML]{1D41D5} 22.60}} & {\color[HTML]{FF0000} 22.97} & \multicolumn{1}{c|}{{\color[HTML]{FF0000} 21.81}} & {\color[HTML]{FF0000} 21.45} & \multicolumn{1}{c|}{{\color[HTML]{FF0000} 20.47}} & {\color[HTML]{FF0000} 20.53} \\
B4 (p=0.7)             & \multicolumn{1}{c|}{{\color[HTML]{1D41D5} 23.29}} & 23.48                        & \multicolumn{1}{c|}{{\color[HTML]{1D41D5} 22.20}} & 22.44                        & \multicolumn{1}{c|}{22.45}                        & 22.98                        & \multicolumn{1}{c|}{21.50}                        & 21.24                        & \multicolumn{1}{c|}{20.23}                        & 20.43                        \\
B8 (p=0.7)             & \multicolumn{1}{c|}{22.84}                        & 23.57                        & \multicolumn{1}{c|}{21.73}                        & 22.46                        & \multicolumn{1}{c|}{21.92}                        & {\color[HTML]{1D41D5} 23.00} & \multicolumn{1}{c|}{20.96}                        & 20.95                        & \multicolumn{1}{c|}{19.88}                        & 20.43                        \\
B12 (p=0.7)            & \multicolumn{1}{c|}{23.03}                        & {\color[HTML]{1D41D5} 23.65} & \multicolumn{1}{c|}{21.89}                        & {\color[HTML]{FF0000} 22.50} & \multicolumn{1}{c|}{22.09}                        & {\color[HTML]{FF0000} 23.04} & \multicolumn{1}{c|}{21.07}                        & 20.94                        & \multicolumn{1}{c|}{20.00}                        & {\color[HTML]{1D41D5} 20.47} \\
B16 (p=0.7)            & \multicolumn{1}{c|}{23.01}                        & 23.59                        & \multicolumn{1}{c|}{21.88}                        & {\color[HTML]{1D41D5} 22.46} & \multicolumn{1}{c|}{22.09}                        & 23.00                        & \multicolumn{1}{c|}{21.05}                        & 20.92                        & \multicolumn{1}{c|}{19.98}                        & 20.42                        \\
quarter (p=0.7)        & \multicolumn{1}{c|}{22.49}                        & 23.17                        & \multicolumn{1}{c|}{21.39}                        & 21.91                        & \multicolumn{1}{c|}{21.60}                        & 22.60                        & \multicolumn{1}{c|}{20.83}                        & 20.63                        & \multicolumn{1}{c|}{19.71}                        & 20.12                        \\
half (p=0.7)           & \multicolumn{1}{c|}{21.76}                        & 22.71                        & \multicolumn{1}{c|}{20.86}                        & 21.75                        & \multicolumn{1}{c|}{21.17}                        & 22.49                        & \multicolumn{1}{c|}{19.84}                        & 19.83                        & \multicolumn{1}{c|}{19.08}                        & 19.70                        \\
all (p=0.7)            & \multicolumn{1}{c|}{21.76}                        & 22.71                        & \multicolumn{1}{c|}{20.86}                        & 21.74                        & \multicolumn{1}{c|}{21.16}                        & 22.48                        & \multicolumn{1}{c|}{19.83}                        & 19.83                        & \multicolumn{1}{c|}{19.07}                        & 19.69                        \\ \hline
                       & \multicolumn{1}{c|}{n+j}                          & b+n+j                        & \multicolumn{1}{c|}{n+j}                          & b+n+j                        & \multicolumn{1}{c|}{n+j}                          & b+n+j                        & \multicolumn{1}{c|}{n+j}                          & b+n+j                        & \multicolumn{1}{c|}{n+j}                          & b+n+j                        \\ \hline
Real-SRResNet (p=0)    & \multicolumn{1}{c|}{23.17}                        & 22.75                        & \multicolumn{1}{c|}{22.01}                        & 21.74                        & \multicolumn{1}{c|}{22.67}                        & 22.39                        & \multicolumn{1}{c|}{21.37}                        & {\color[HTML]{1D41D5} 20.82} & \multicolumn{1}{c|}{20.41}                        & 20.09                        \\
last (p=0.7)           & \multicolumn{1}{c|}{{\color[HTML]{FF0000} 23.53}} & {\color[HTML]{FF0000} 23.04} & \multicolumn{1}{c|}{{\color[HTML]{3531FF} 22.26}} & {\color[HTML]{1D41D5} 21.97} & \multicolumn{1}{c|}{{\color[HTML]{1D41D5} 22.81}} & {\color[HTML]{FF0000} 22.51} & \multicolumn{1}{c|}{{\color[HTML]{FF0000} 21.65}} & {\color[HTML]{FF0000} 21.03} & \multicolumn{1}{c|}{{\color[HTML]{1D41D5} 20.63}} & {\color[HTML]{FF0000} 20.22} \\
B4 (p=0.7)             & \multicolumn{1}{c|}{23.50}                        & {\color[HTML]{1D41D5} 22.95} & \multicolumn{1}{c|}{{\color[HTML]{FF0000} 22.39}} & {\color[HTML]{FF0000} 21.98} & \multicolumn{1}{c|}{{\color[HTML]{FF0000} 22.85}} & {\color[HTML]{1D41D5} 22.46} & \multicolumn{1}{c|}{{\color[HTML]{1D41D5} 21.42}} & 20.79                        & \multicolumn{1}{c|}{20.61}                        & {\color[HTML]{1D41D5} 20.09} \\
B8 (p=0.7)             & \multicolumn{1}{c|}{23.52}                        & 22.83                        & \multicolumn{1}{c|}{22.33}                        & 21.84                        & \multicolumn{1}{c|}{22.75}                        & 22.32                        & \multicolumn{1}{c|}{21.17}                        & 20.42                        & \multicolumn{1}{c|}{20.59}                        & 19.97                        \\
B12 (p=0.7)            & \multicolumn{1}{c|}{{\color[HTML]{FF0000} 23.59}} & 22.91                        & \multicolumn{1}{c|}{{\color[HTML]{1D41D5} 22.37}} & 21.90                        & \multicolumn{1}{c|}{22.79}                        & 22.37                        & \multicolumn{1}{c|}{21.22}                        & 20.47                        & \multicolumn{1}{c|}{{\color[HTML]{FF0000} 20.64}} & 20.03                        \\
B16 (p=0.7)            & \multicolumn{1}{c|}{{\color[HTML]{1D41D5} 23.57}} & 22.86                        & \multicolumn{1}{c|}{22.34}                        & 21.86                        & \multicolumn{1}{c|}{22.74}                        & 22.32                        & \multicolumn{1}{c|}{21.16}                        & 20.43                        & \multicolumn{1}{c|}{20.60}                        & 19.98                        \\
quarter (p=0.7)        & \multicolumn{1}{c|}{22.97}                        & 22.68                        & \multicolumn{1}{c|}{21.85}                        & 21.64                        & \multicolumn{1}{c|}{22.46}                        & 22.21                        & \multicolumn{1}{c|}{20.77}                        & 20.33                        & \multicolumn{1}{c|}{20.21}                        & 19.89                        \\
half (p=0.7)           & \multicolumn{1}{c|}{22.58}                        & 21.92                        & \multicolumn{1}{c|}{21.64}                        & 21.12                        & \multicolumn{1}{c|}{22.14}                        & 21.73                        & \multicolumn{1}{c|}{20.02}                        & 19.38                        & \multicolumn{1}{c|}{19.88}                        & 19.28                        \\
all (p=0.7)            & \multicolumn{1}{c|}{22.58}                        & 21.92                        & \multicolumn{1}{c|}{21.63}                        & 21.12                        & \multicolumn{1}{c|}{22.13}                        & 21.72                        & \multicolumn{1}{c|}{20.01}                        & 19.37                        & \multicolumn{1}{c|}{19.87}                        & 19.28                        \\ \hline
\end{tabular}
      \end{center}
      \vskip -0.3cm
      \caption{The PSNR ($\db$) results of Real-SRResNet with different dropout positions. Each of two columns gives a test set with 8 types of degradations. We apply bicubic, blur, noise and jpeg to generate the degradation, e.g. clean means only bicubic, noise means bicubic $\to$ noise, b+n+j means blur $\to$ bicubic $\to$ noise $\to$ jpeg. {\color[HTML]{FF0000}Red}/{\color[HTML]{0000FF}Blue} text: best/second-best PSNR ($\db$).}
      \vskip -0.3cm
      \label{table:positions}
      \end{table*}

\paragraph{Probabilities.}
We show the performance difference of using different dropout probabilities in \tablename~\ref{table:probabilities}. 
The results of Real-SRResNet with dropout probabilities from 10\% to 90\% are better than the results without dropout.
We select $p=0.7$ for Real-SRResNet and $p=0.5$ for Real-RRDB. 
Nevertheless, other dropout probabilities are also useful.
These results demonstrate that dropout methods can improve the generalization ability of SR networks stably.

\section{Training Curves of Models}
\label{sec:Training Curves of Models}
    \begin{figure}[ht!]
      \centering
      \includegraphics[width=1.0\linewidth{}]{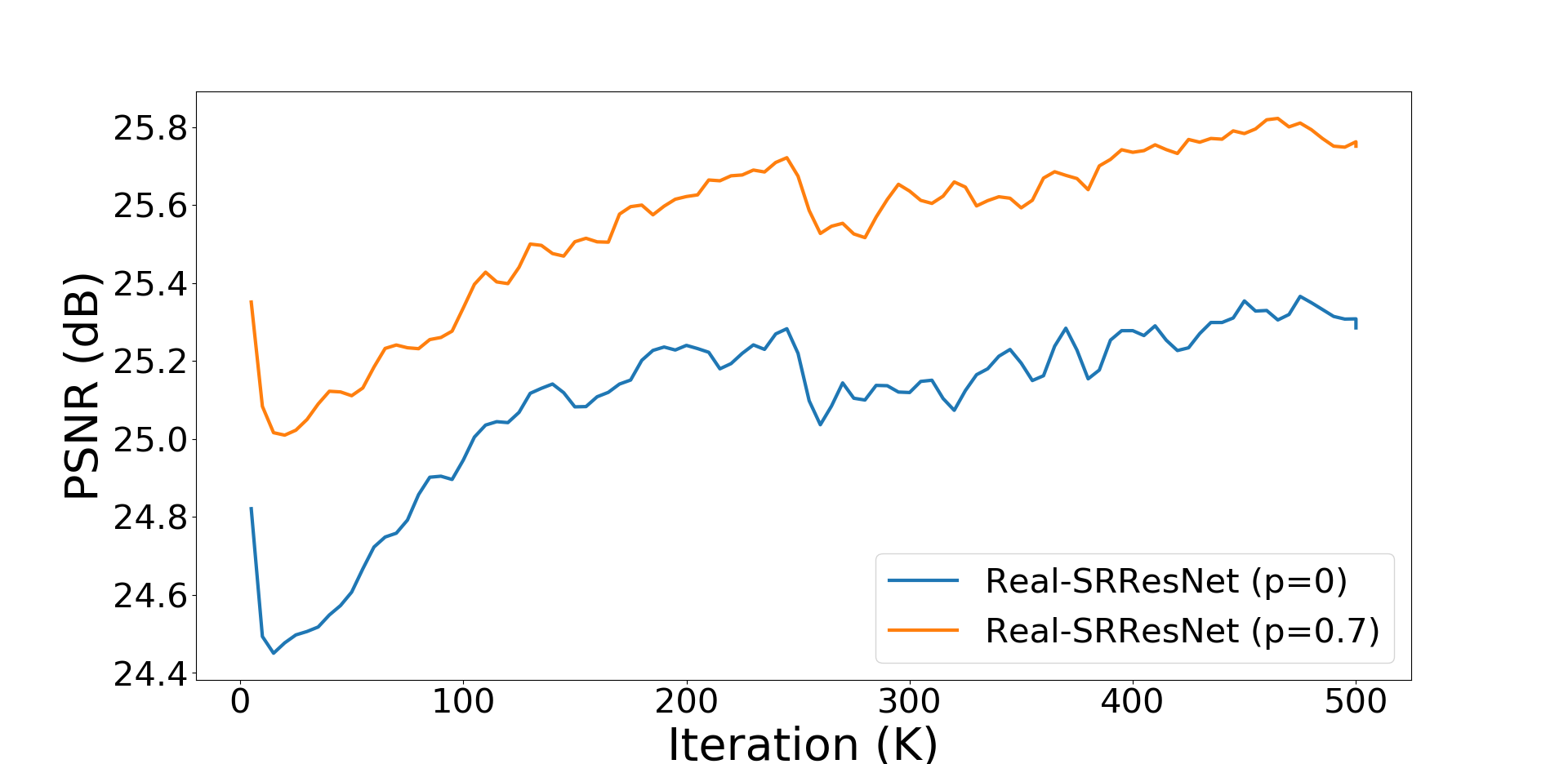}
      \vskip -0.1cm
      \caption{
      Training curves of Real-SRResNet. The validation set is Set5~\cite{Set5} (clean).}
      \label{fig:SRResNet}
      \vskip -0.7cm
  \end{figure}

  \begin{figure}[ht!]
      \includegraphics[width=1.0\linewidth]{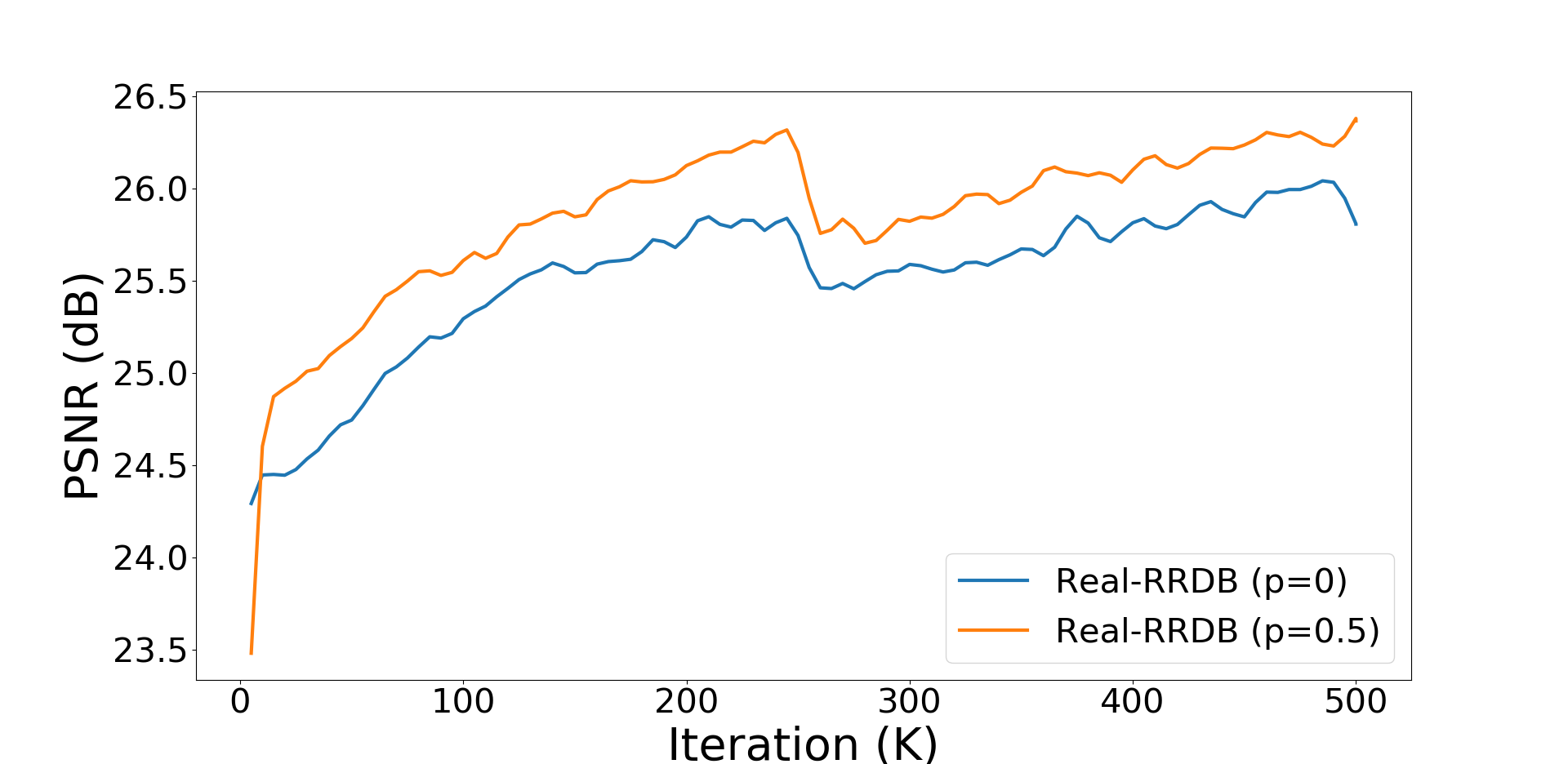}
      \vskip -0.1cm
      \caption{
      Training curves of Real-RRDB. The validation set is Set5~\cite{Set5} (clean).}
      \label{fig:RRDB}
      \vskip -0.3cm
  \end{figure}
  \vskip -0.7cm
  \begin{figure}[ht!]
      \includegraphics[width=1.0\linewidth]{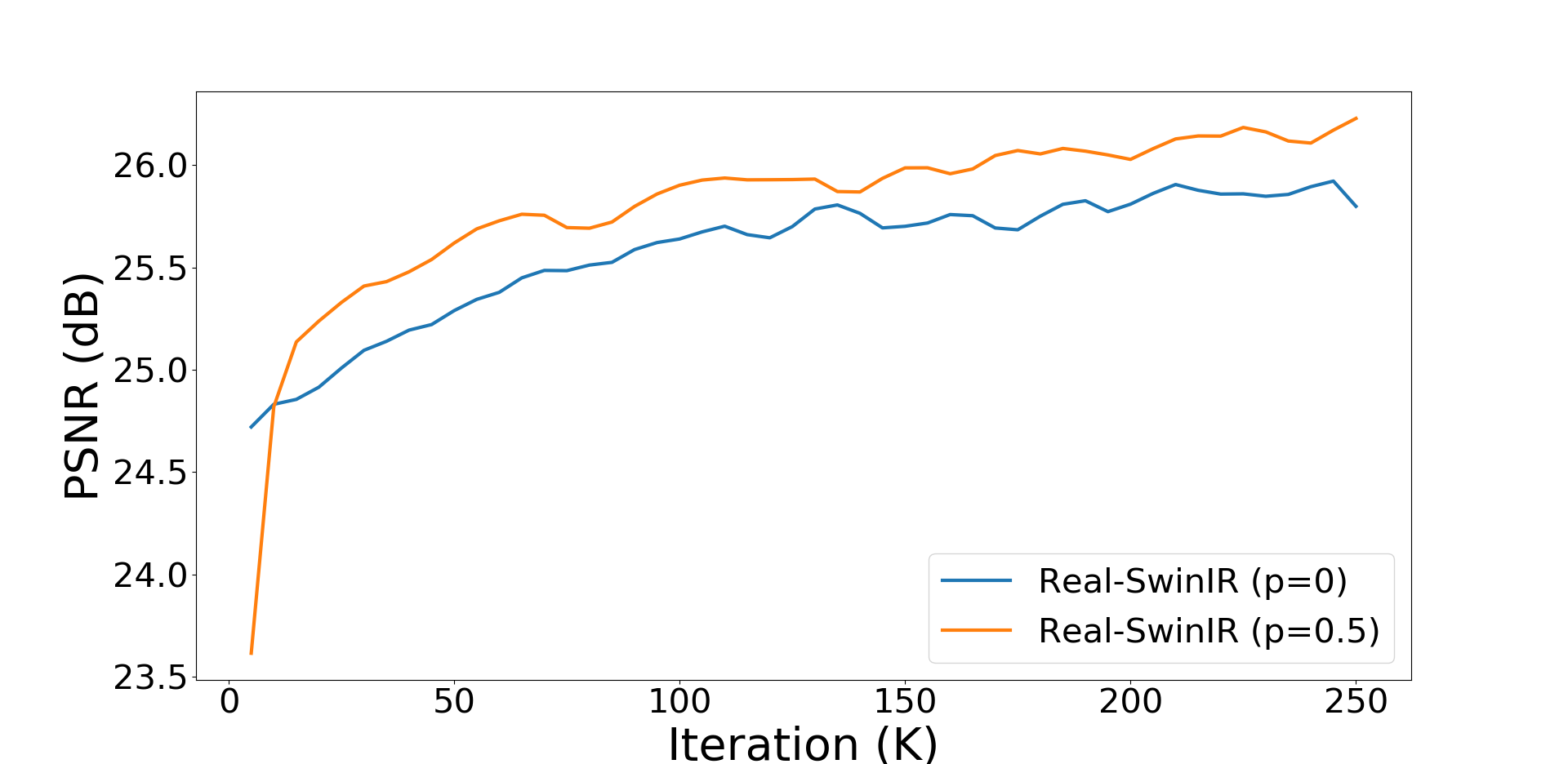}
      \vskip -0.1cm
      \caption{
      Training curves of Real-SwinIR. The validation set is Set5~\cite{Set5} (clean).}
      \label{fig:SwinIR}
  \end{figure}

Is the improvement in performance on account of dropout changes the convergence characteristics of networks?
We visualize the training curves of Real-SRResNet (\figurename~\ref{fig:SRResNet}), Real-RRDB (\figurename~\ref{fig:RRDB}) and SwinIR(\figurename~\ref{fig:SwinIR}).
As shown in \figurename~\ref{fig:SRResNet},\ref{fig:RRDB} and \ref{fig:SwinIR}, dropout does not change the convergence characteristics of the networks.
During the training process, a PSNR comparison of Set5 (clean) shows that the models (both SRResNet, RRDB and SwinIR) with dropout consistently perform better than the normal models. 
However, they have convergence curves that are almost exactly the same.

\section{More Qualitative Results}
\label{sec:More Qualitative Results}

\begin{figure*}[t!]
    \centering
    \includegraphics[width=1.0\linewidth]{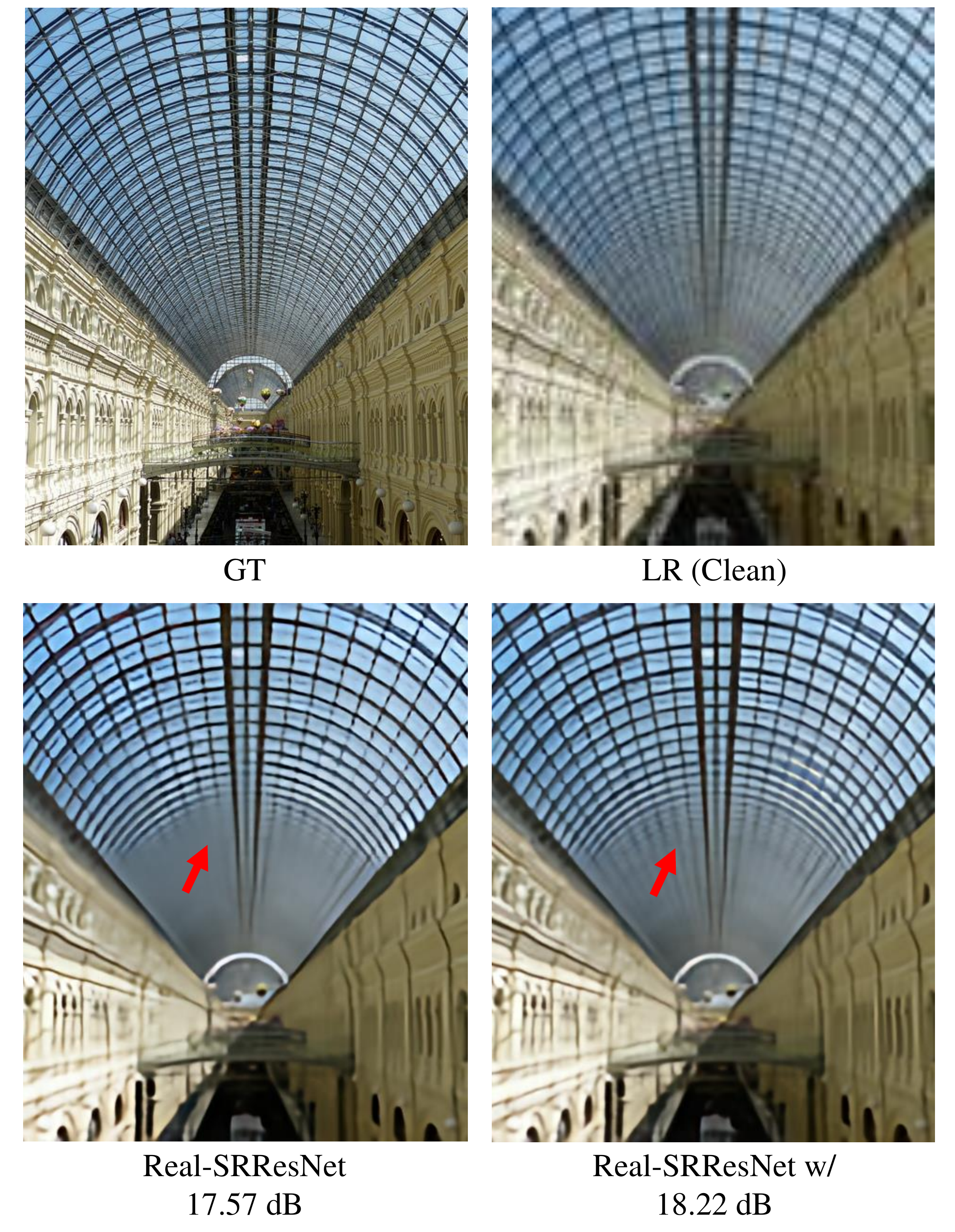}
    \caption{Visual results of ``Clean''. We use ``w/'' to represent the model with dropout. (Zoom in for best view)}
    \label{fig:1}
\end{figure*}

\begin{figure*}[t!]
    \centering
    \includegraphics[width=1.0\linewidth]{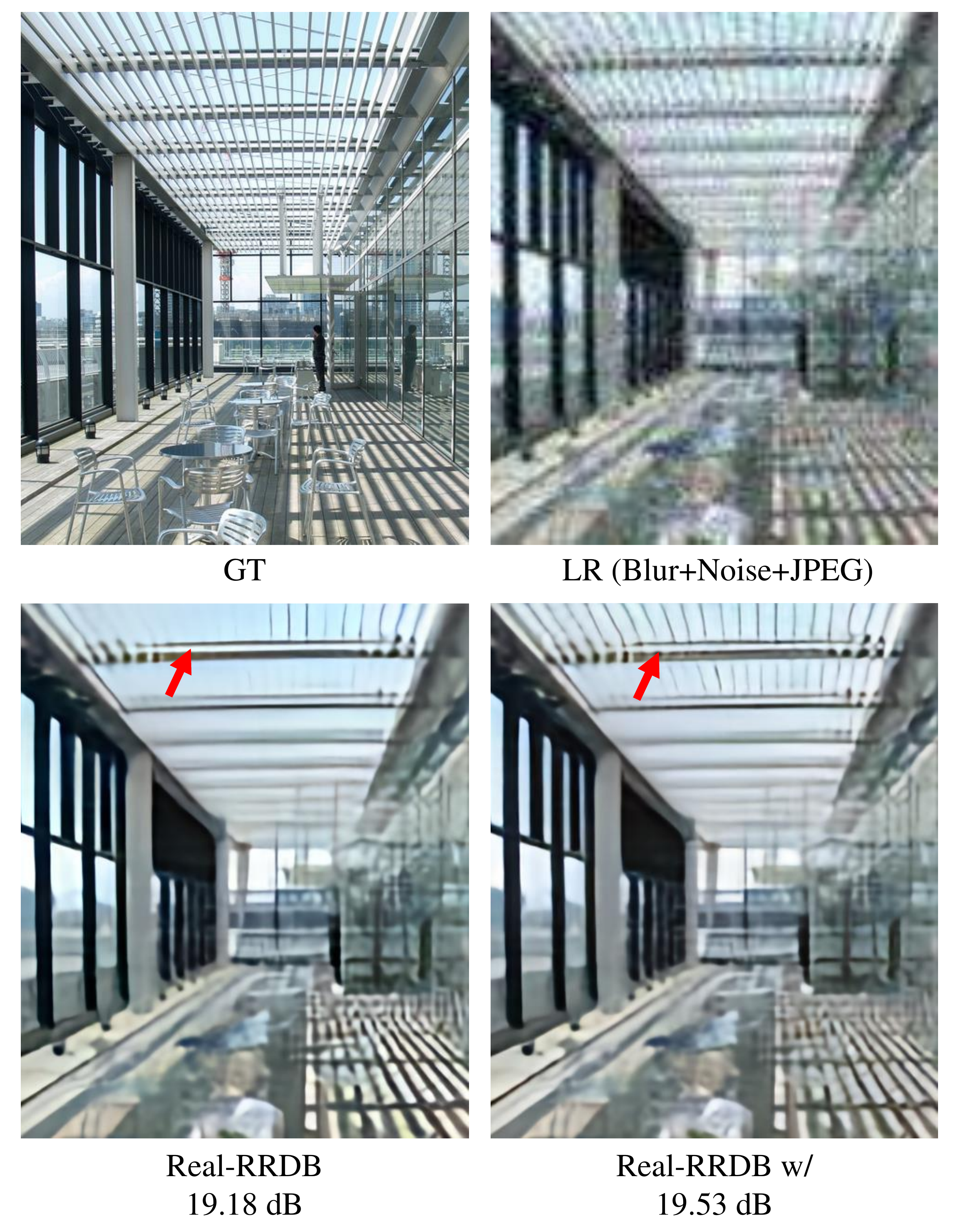}
    \caption{Visual results of ``Blur+Noise+JPEG''. We use ``w/'' to represent the model with dropout. (Zoom in for best view)}
    \label{fig:8}
\end{figure*}

In this section, we provide additional qualitative results on different degradations to clearly show the effectiveness of dropout (see \figurename~\ref{fig:1} to \figurename~\ref{fig:8}). Following the testing setting, we select Gaussian blur with kernel size 21 and standard deviation 2 (denoted by ``Blur''), bicubic downsampling (denoted by ``Clean''), Gaussian noise with a standard deviation 20 (denoted by ``Noise'') and JPEG compression with quality 50 (denoted by ``JPEG'') as degradations to show the qualitative results.
We also include complex mixed degradations that are combined by the above component.
For these mixed degradations, we synthesize them in the same order as the training method.

\begin{figure*}[t!]
    \centering
    \includegraphics[width=1.0\linewidth]{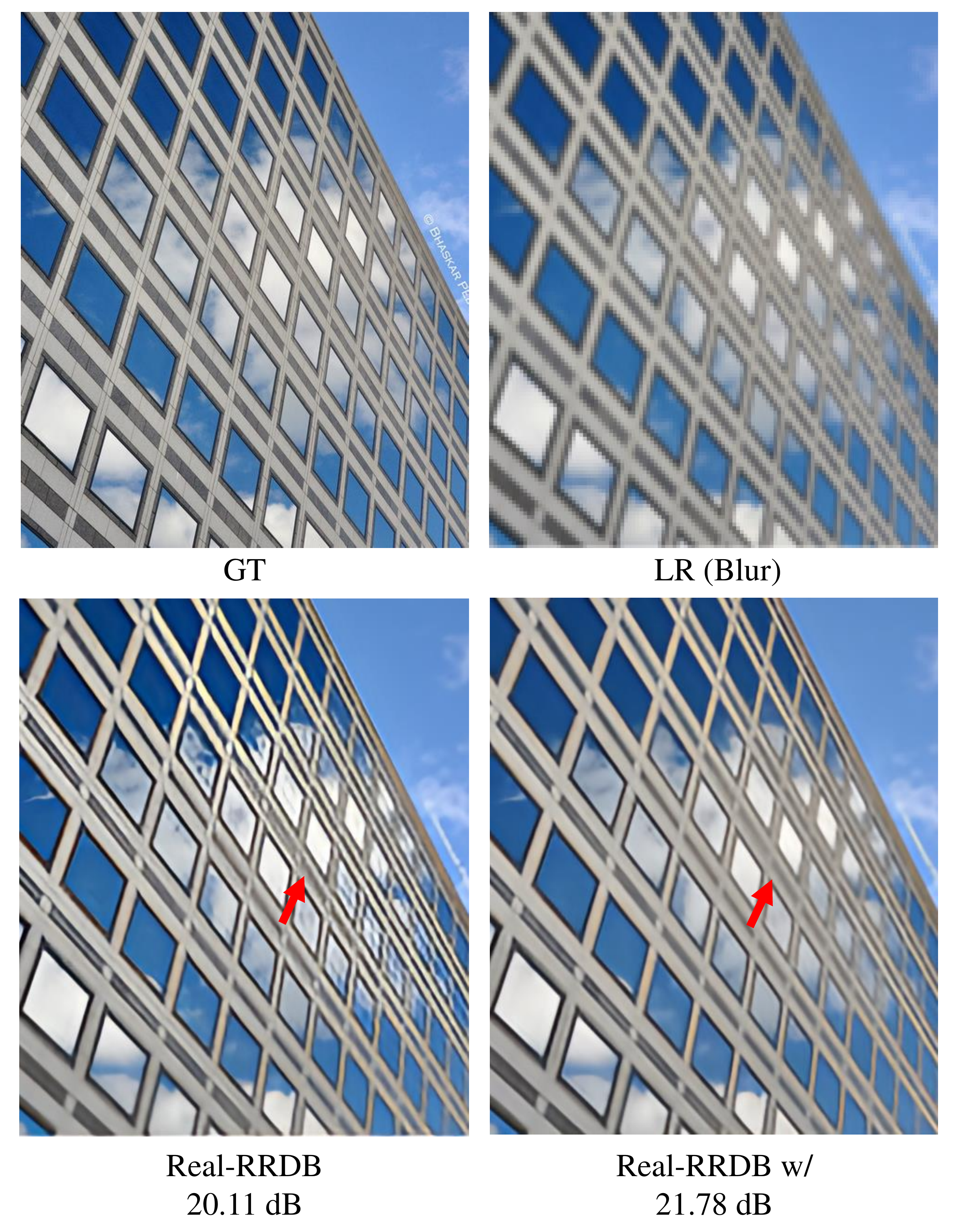}
    \caption{Visual results of ``Blur''. We use ``w/'' to represent the model with dropout. (Zoom in for best view)}
    \label{fig:2}
\end{figure*}

\begin{figure*}[t!]
    \centering
    \includegraphics[width=1.0\linewidth]{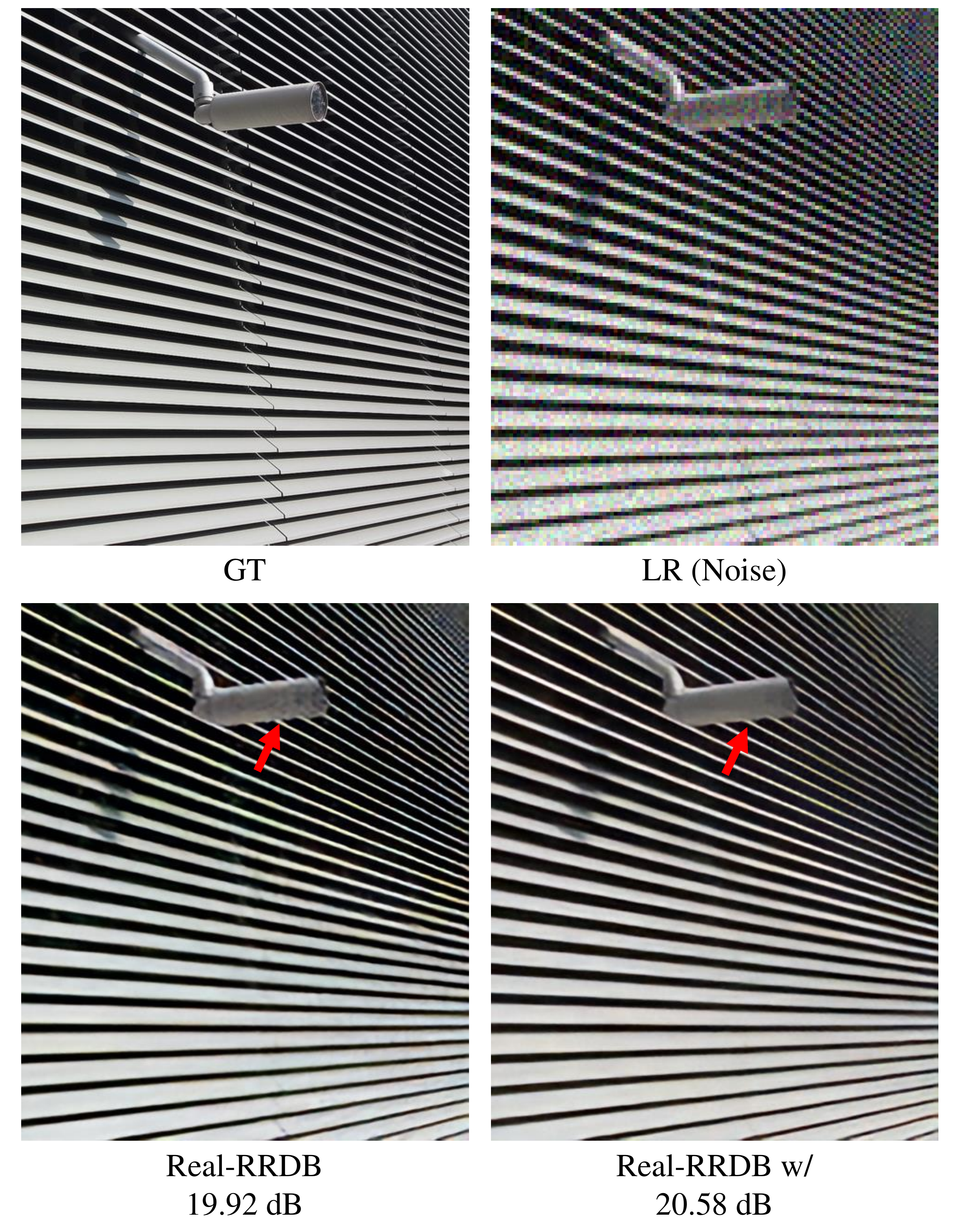}
    \caption{Visual results of ``Noise''. We use ``w/'' to represent the model with dropout. (Zoom in for best view)}
    \label{fig:3}
\end{figure*}

\begin{figure*}[t!]
    \centering
    \includegraphics[width=1.0\linewidth]{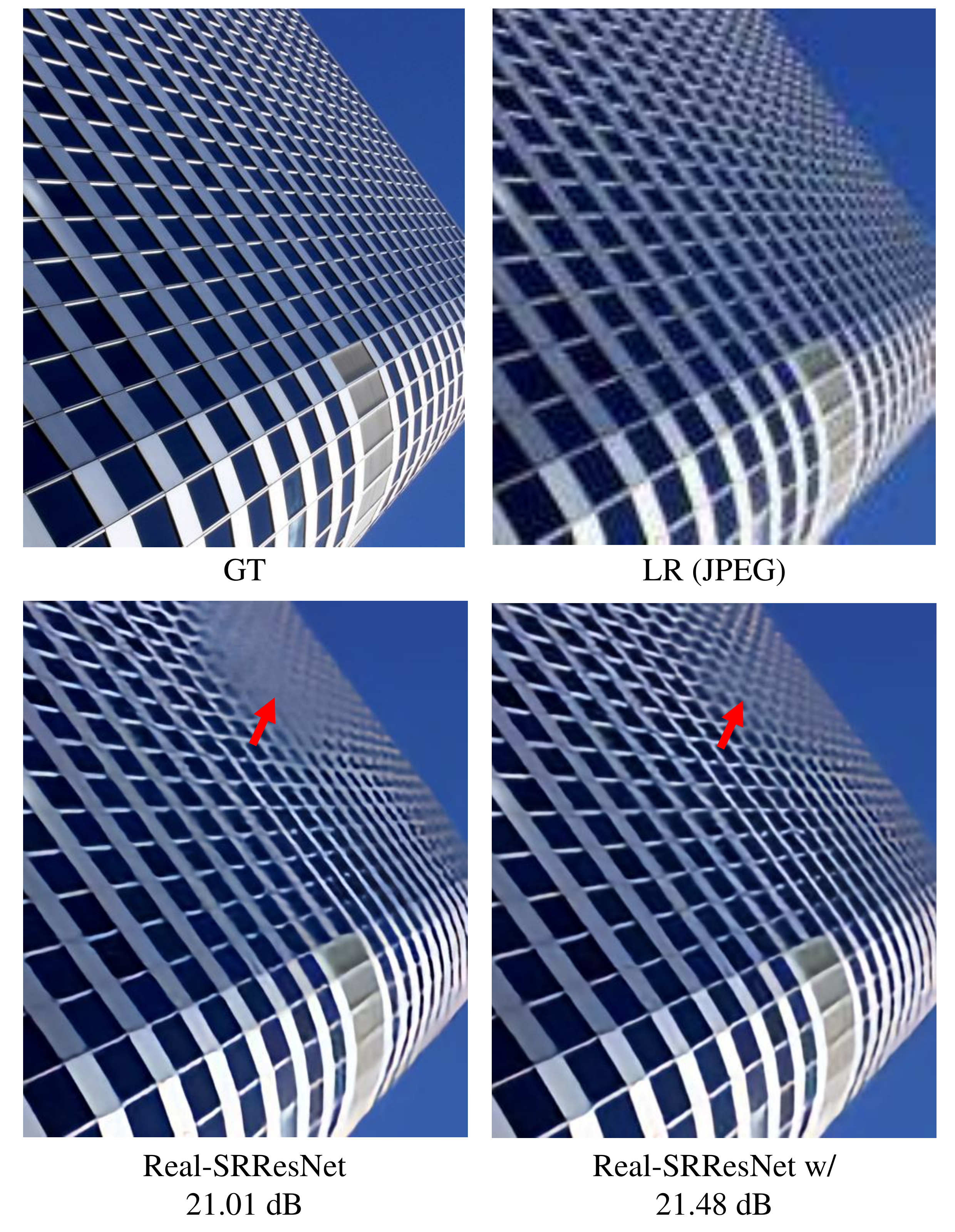}
    \caption{Visual results of ``JPEG''. We use ``w/'' to represent the model with dropout. (Zoom in for best view)}
    \label{fig:4}
\end{figure*}

\begin{figure*}[t!]
    \centering
    \includegraphics[width=1.0\linewidth]{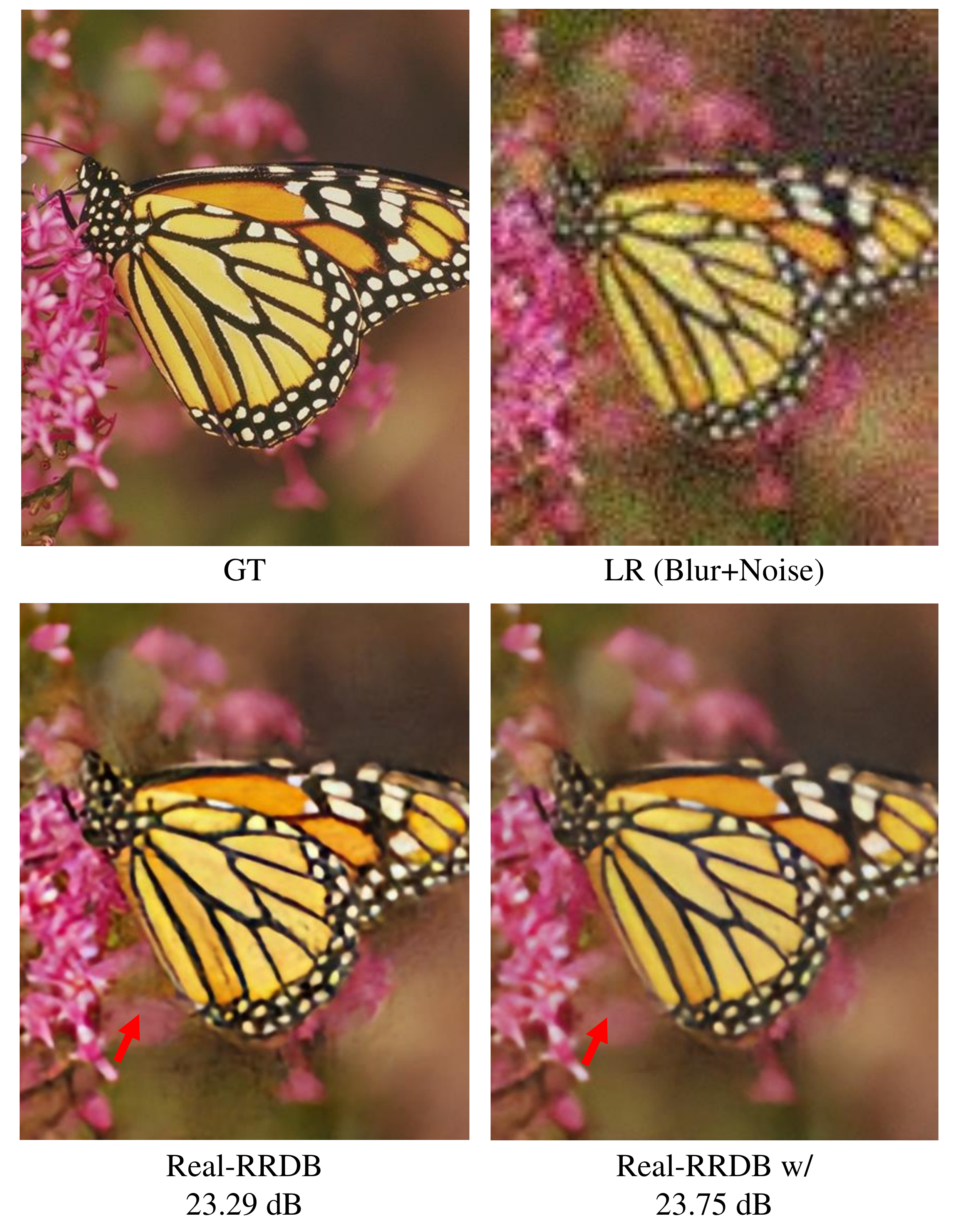}
    \caption{Visual results of ``Blur+Noise''. We use ``w/'' to represent the model with dropout. (Zoom in for best view)}
    \label{fig:5}
\end{figure*}

\begin{figure*}[t!]
    \centering
    \includegraphics[width=1.0\linewidth]{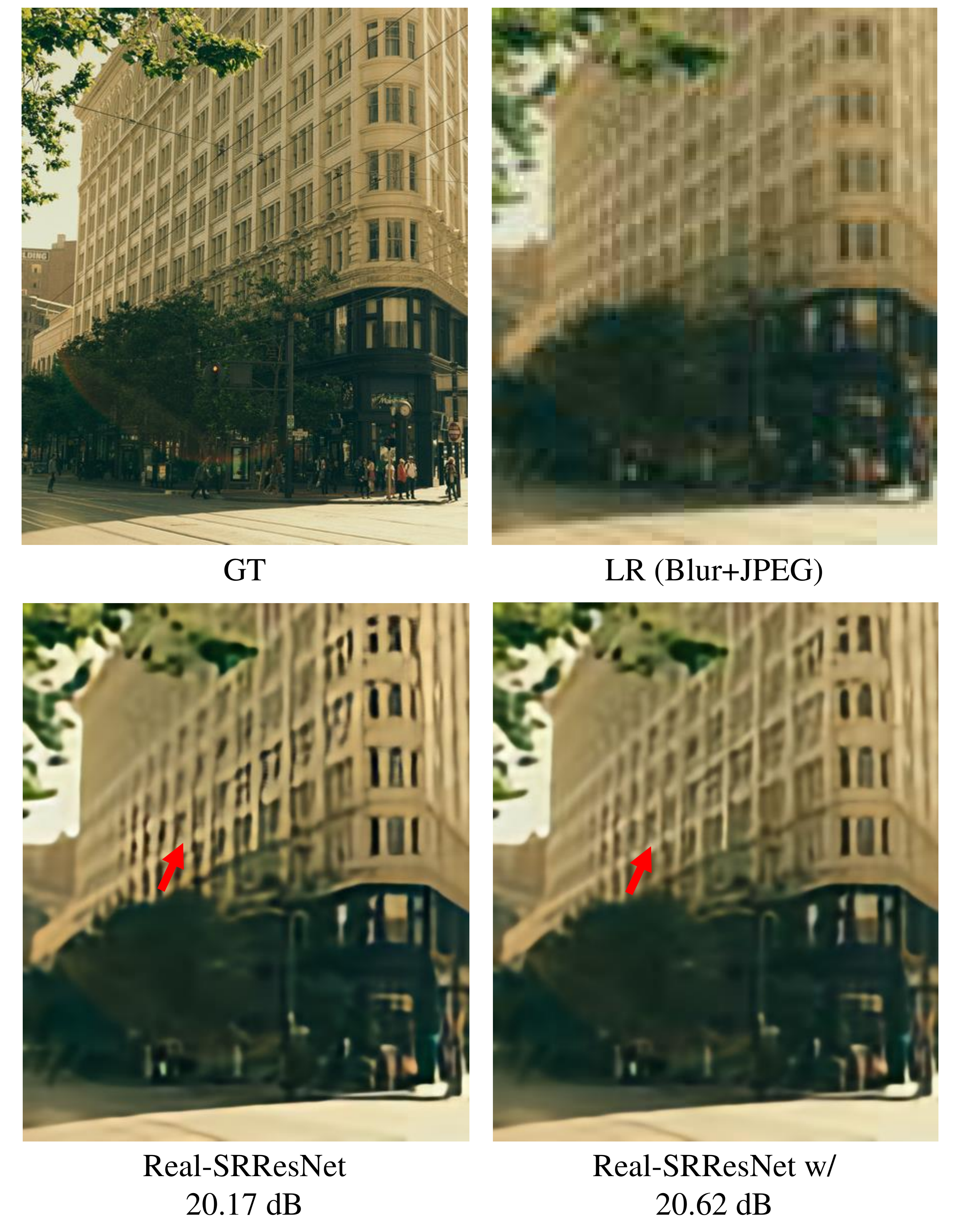}
    \caption{Visual results of ``Blur+JPEG''. We use ``w/'' to represent the model with dropout. (Zoom in for best view)}
    \label{fig:6}
\end{figure*}

\begin{figure*}[t!]
    \centering
    \includegraphics[width=1.0\linewidth]{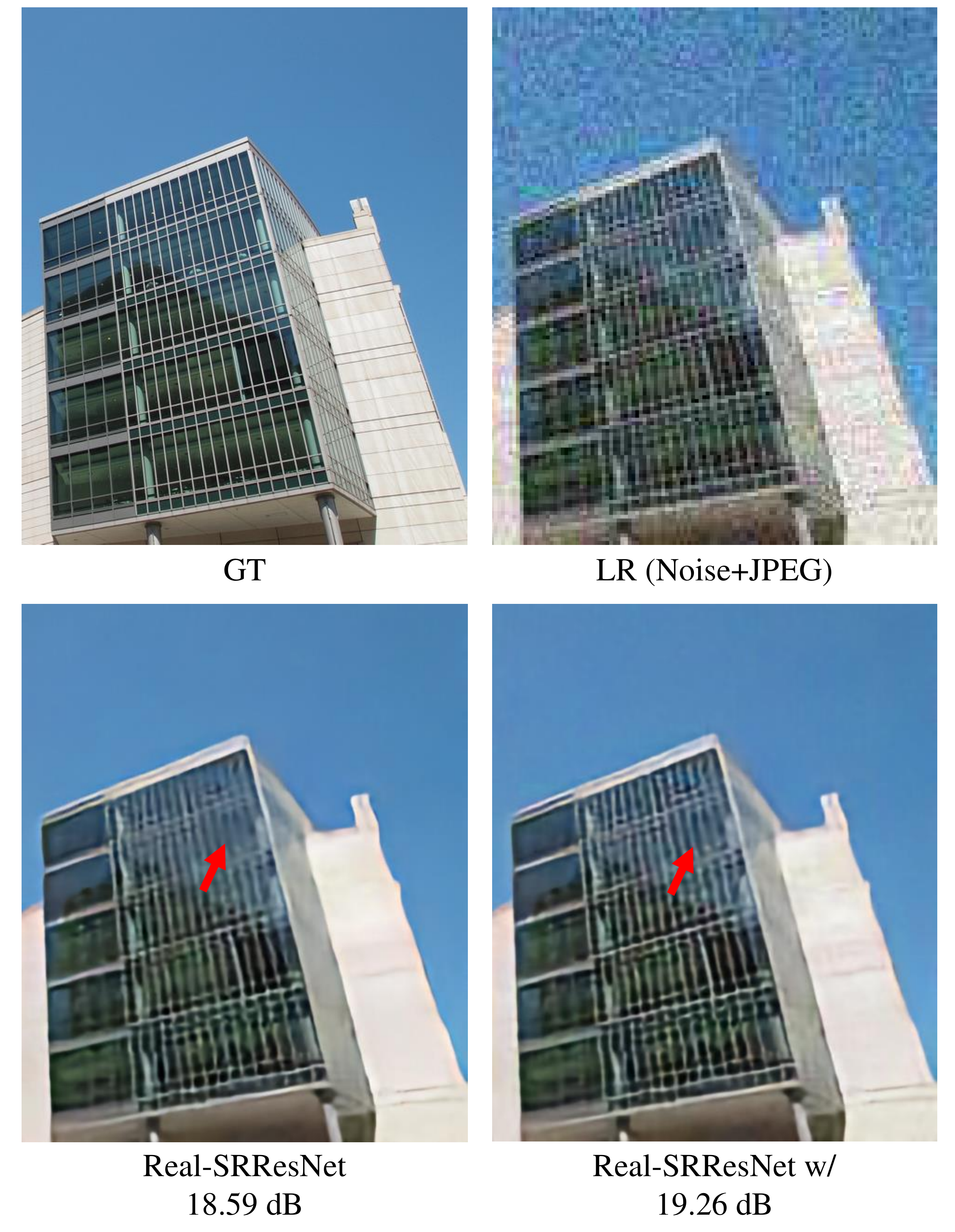}
    \caption{Visual results of ``Noise+JPEG''. We use ``w/'' to represent the model with dropout. (Zoom in for best view)}
    \label{fig:7}
\end{figure*}

\end{document}